\documentclass[letterpaper, 10 pt, conference]{ieeeconf}  % Comment this line out if you need a4paper

\IEEEoverridecommandlockouts                              % This command is only needed if 
\usepackage{graphics} % for pdf, bitmapped graphics files
\usepackage{graphicx} % for pdf, bitmapped graphics files
\makeatletter
\let\MYcaption\@makecaption
\makeatother

\usepackage[font=scriptsize]{subcaption}

\makeatletter
\let\@makecaption\MYcaption
\makeatother

\usepackage{amsmath} % assumes amsmath package installed
\usepackage{amssymb}  % assumes amsmath package installed
\usepackage{algorithm, algorithmic} % for writing algorithms
\usepackage{bm}
\usepackage{xcolor}
\usepackage{dblfloatfix}
\usepackage{multirow}
\usepackage{cite} % For compressed ranges (works with IEEETran)
\usepackage[textsize=tiny]{todonotes}

\usepackage{amsthm}

\newtheorem{proposition}{Proposition}

\newtheorem{corollary}{Corollary}
\newtheorem{specialcase}{Special Case}

\let\labelindent\relax
\usepackage[shortlabels]{enumitem}

\newcommand{\tflag}[1]{}
\newcommand{\tfflag}[1]{}

\newcommand{\blue}[1]{% Make the text original color
	{#1}% 
}

\newcommand{\blindreview}[1]{}
\newcommand{\original}[1]{{#1}}

\renewcommand{\baselinestretch}{0.98}

\makeatletter
\let\NAT@parse\undefined
\makeatother
\usepackage{hyperref}  %hyperref still needs to be put at the end!
\setlist[itemize]{noitemsep, topsep=0pt}

\title{\LARGE \bf
Uncertainty-Aware Planning for Heterogeneous Robot Teams using Dynamic Topological Graphs and Mixed-Integer Programming
}

\author{
Cora A. Duggan$^{1,2}$, 
Kevin C. Wolfe$^{1}$,
Bradley Woosley$^{3}$,
Marin Kobilarov$^{2}$,
and Joseph Moore$^{1,2}$% <-this % stops a space
\thanks{$^{1}$Johns Hopkins University Applied Physics Laboratory, Laurel, MD
	20723, USA. Email: {\tt\small Cora.Duggan@jhuapl.edu}}%
\thanks{$^{2}$Department of Mechanical Engineering, Johns Hopkins University, Baltimore, MD 21218, USA. Email: {\tt\small jlmoore@jhu.edu}}%
\thanks{$^{3}$DEVCOM Army Research Laboratory, Adelphi, MD 20783, USA.}%
\thanks{\noindent Distribution statement A. Approved for public release; distribution is unlimited.}%
}

\begin{document}

\maketitle
\thispagestyle{empty}
\pagestyle{empty}

%%%%%%%%%%%%%%%%%%%%%%%%%%%%%%%%%%%%%%%%%%%%%%%%%%%%%%%%%%%%%%%%%%%%%%%%%%%%%%%%
\begin{abstract}
Multi-robot planning and coordination in uncertain environments is a fundamental computational challenge, since the belief space increases exponentially with the number of robots. 
In this paper, we address the problem of planning in uncertain environments with a heterogeneous robot team of fast scout vehicles for information gathering and more risk-averse carrier robots from which the scouts vehicles are deployed. \tflag{R-2-2}\blue{To overcome the computational challenges, %associated with multi-robot planning in the presence of environmental uncertainty, 
we represent the environment and operational scenario using a topological graph, where the parameters of the edge weight distributions vary with the state of the robot team on the graph, 
% While this belief space representation still scales exponentially with the number of robots, we formulate a computationally efficient mixed-integer program which is capable of generating optimal multi-robot plans in seconds. 
%Additionally, to enable planning when the belief space scales exponentially with the number of robots, 
and we formulate a computationally efficient mixed-integer program which removes the dependence on the number of robots from its decision space. 
%Overall, our formulation enables planning when the belief space scales exponentially with the number of robots. 
Our formulation results in the capability to generate optimal multi-robot, long-horizon plans in seconds that could otherwise be computationally intractable. \tflag{R-1-4}Ultimately our approach enables real-time re-planning, since the computation time is significantly faster than the time to execute one step.}
We evaluate our algorithm in a scenario where the robot team must traverse an environment while minimizing detection by observers in positions that are uncertain to the robot team. We demonstrate that our method is computationally tractable, %for real-time re-planning in changing environments, 
can improve performance in the presence of imperfect information, and can be adjusted for different risk profiles.

%Multi-robot planning and coordination in uncertain environments is a fundamental challenge. Oftentimes, such planning problems are computationally intractable, since the belief space increases exponentially with the number of robots. In this paper, we address the problem of planning in uncertain environments with a heterogeneous robot team whose members possess different risk profiles. In particular, we consider the case where the robot team is comprised of fast scout vehicles capable of gathering information about the environment and risk-averse carrier robots from which the scout vehicles are deployed. Instead of planning in the continuous belief space, we represent our environment as a topological graph with uncertain edge weights, where the edge weight distributions vary with the state of the robot team on the graph. While this belief space representation still scales exponentially with the number of robots, we formulate a computationally efficient mixed-integer program which is capable of generating optimal multi-robot plans in seconds. We evaluate our approach in a representative scenario where the robot team must move through an environment while minimizing detection by observers in positions that are uncertain to the robot team. We demonstrate that our approach is sufficiently computationally tractable for real-time re-planning in changing environments, can improve performance in the presence of imperfect information, and can be adjusted to accommodate different risk profiles.
\end{abstract}

%%%%%%%%%%%%%%%%%%%%%%%%%%%%%%%%%%%%%%%%%%%%%%%%%%%%%%%%%%%%%%%%%%%%%%%%%%%%%%%%
\section{INTRODUCTION}

As multi-robot systems are deployed in real-world scenarios, capabilities for reasoning about environmental uncertainty become essential. For instance, to be successful in uncertain hazardous environments, a robot team must balance achieving mission objectives (e.g., %search or 
navigating to a goal) with avoiding unforeseen hazards that could catastrophically impact team performance. However, achieving unified multi-robot planning and coordination in uncertain environments poses a significant computational challenge, since the belief space \tflag{R-2-3}\blue{(i.e., the space of all possible probability distributions over the state space)} grows exponentially with the number of robots \blue{due to accounting for robot interactions}. 
\begin{figure}[t]
	\vspace*{2mm}
	\centering
	\includegraphics[width=0.98\columnwidth]{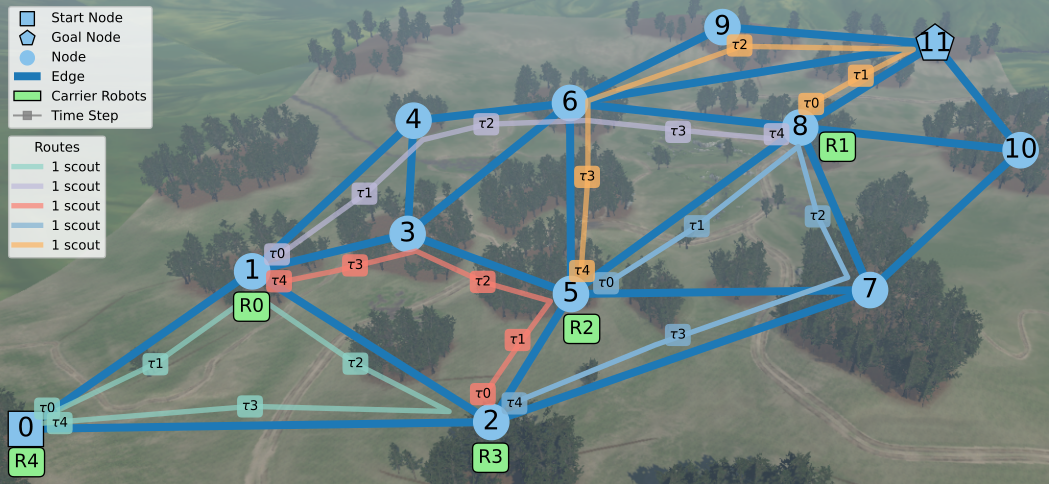}
    \vspace*{-2mm}
	\caption{A dynamic topological graph, in a meadow environment from \cite{NatureManufacture} with uncertain edge weights, applied to a reconnaissance test scenario. The routes through the graph show the paths of scout agents deployed from carrier vehicles to explore the environment. 
    % The nodes are in forested regions of cover. Ten robot scouts start at node 1 with the goal of at least one robot reaching node 2, across the meadow. 
    % The edges of the graph have cost for transitioning between nodes that is a function of their distance, detectability, and vulnerability.  
    % This cost can be reduced by moving in teams and by providing overwatch, where robots at a particular node oversee the movement of robots along a corresponding edge to help mitigate some of the risk of traversing. Overwatch opportunities are indicated with an arrow from the overwatch node pointing toward the edge that can be monitored. 
    % The robot team routes through the graph, solved for with our proposed method, represent a solution to this problem.
	}
	\label{fig:meadow_example}
	\vspace*{-7mm}
\end{figure}

In this paper, we present a computationally tractable approach for planning and coordination of heterogeneous robot teams operating in uncertain environments when the robot team members possess different risk profiles. In particular, we consider the case where the multi-robot team consists of fast, risk-tolerant scout vehicles and risk-averse carrier vehicles. 
% Rather than solving the multi-robot planning problem in a continuous belief space, we segment the environment based on the operational scenario and construct a topological graph, as seen in Fig.~\ref{fig:meadow_example}. 
\tflag{R-2-4}\blue{To create a computationally tractable decision space, we construct a dynamic topological graph structure, as seen in Fig.~\ref{fig:meadow_example}, by discretizing the environment based on the operational scenario and \textit{a priori} terrain data. The graph also \tflag{R-2-5} embeds the critical relationship between environmental uncertainty and the state of the robot team in its stochastic edge weights.} We then use this graph to formulate an optimization problem with Mixed-Integer Programming (MIP) for generating multi-robot plans that reason about uncertainty while satisfying spatial and temporal coordination constraints.

To evaluate our approach, we consider a scenario in which a heterogeneous robot team must minimize detection \tflag{R-2-6}\blue{by observers} while navigating through an environment where observer positions are dynamic and not fully known. Our numerical results demonstrate that our approach is computationally tractable for medium-sized robot teams ($\sim$ 10 agents) operating in large-scale environments and can enable adaptive re-planning as new information is received. We also detail the advantages of our approach through an ablation study that shows the value of reasoning about uncertainty, leveraging information-gathering scout robots, and including certainty decay. 
\original{Unlike our prior approach}\blindreview{Unlike the approach} 
in \cite{dimmig2023mip}, the method presented in this paper can reason about environmental uncertainty, utilize heterogeneous risk profiles, and leverage rapid re-planning to adapt to new information. 
Our contributions are as follows: 
\begin{itemize}
\item A novel dynamic topological graph formulation for embedding environmental uncertainty.
\item A compact mixed-integer optimization problem capable of encoding graph edge uncertainty and heterogeneous teaming constraints.
\item A computationally efficient approach for graph-planning that can enable online adaptation to new information and accommodate different risk profiles of the team.
\end{itemize}

%%%%%%%%%%%%%%%%%%%%%%%%%%%%%%%%%%%%%%%%%%%%%%%%%%%%%%%%%%%%%%%%%%%%%%%%%%%%%%%%
\section{RELATED WORK}
Multi-robot planning and coordination in uncertain environments is often posed as a multi-agent Partially Observable Markov Decision Process (POMDP)\cite{Zhang2021}. However, %it is well known that 
even for single robot planning problems, generating an optimal solution for a POMDP is computationally intractable \cite{Papadimitriou1987}. 
These computational challenges are further exacerbated for multi-robot systems, 
% In the case of multi-robot systems, these computational challenges are further exacerbated, 
since the belief space increases exponentially with the number of robots \cite{Capitan2013}. To overcome computational limitations, researchers have employed a number of approximations and abstractions to generate sub-optimal solutions. In some cases, researchers have maintained continuous state and action representations and have relied on generating local policies in belief-space to enable multi-robot planning in unknown environments \cite{Indelman2018}. In most cases, however, researchers have relied on a discretization of the state and action space to facilitate computational tractability. For instance, many approaches decompose an overall robot team objective (e.g., exploration) into discrete sub-tasks. Task assignment and planning are often decoupled, where individual robots execute local policies while auction-based methods allocate tasks \cite{Mataric2003,Woosley2021}. In \cite{Capitan2013}, the authors use auctions to allocate POMDP policies for individual robots. Other researchers discretize the action space via macro-actions \cite{Liu2017}.

Graphs have served as an effective means of discretizing state and action spaces to enable planning for single and multi-robot scenarios, including multi-agent path finding \cite{Ma2022} and more complex coordination tasks \cite{dimmig2023mip, Limbu2023}. Oftentimes, nodes represent robot states while edges represent possible transitions between those states. Characteristics such as spatial or temporal relationships between tasks or robots can be embedded into the graph through properties like node connectivity and edge weights. Graphs have also proven to be useful for planning in uncertain environments, with researchers modeling uncertain edge weights as binary \cite{Papadimitriou1991} or continuous random variables \cite{Loui1983,Polychronopoulos1996}, whose distribution could be dependent on decisions made at a prior node \cite{Bertsekas1991}.

To generate plans on these %uncertain 
graphs, researchers have explored search-based, learning-based, and optimization-based approaches. Most search-based approaches have been developed for single robots (e.g., \cite{Chung2019}), while some conflict-based search approaches have been developed for multi-robot systems to address task completion uncertainty or stochastic travel times \cite{Choudhury2022,Peltzer2020}. To overcome challenges associated with problem scale, % (e.g., due to number of robots or size of graph), 
machine learning methods have been applied. In \cite{Omidshafiei2017,Lauri2020}, researchers learn policy graphs to coordinate macro actions to solve a decentralized POMDP. In \cite{Bhattacharya2023}, a graph structure is used with approximate policy iteration for a multi-robot repair problem. Recently, Graph Neural Networks (GNNs) have been investigated toward %demonstrated success %with regards to achieving 
multi-robot coordination in uncertain environments \cite{Tzes2023,Zhang2022}. Advances in optimization, specifically MIP, have also enabled single robot \cite{Buchholz2023} and multi-robot \cite{Okubo2023,Asfora2020} planning in unknown environments using a graph structure, but all report computational challenges. %For single robot systems, \cite{Buchholz2023} considers MIP, but encounters long solve times. %\cite{Okubo2023,Asfora2020} explore MIP-based approaches for multi-robot exploration on graphs, and also report computational challenges.

Although most of the research on the planning and coordination of multi-robot teams under uncertainty has explored homogeneous teams, researchers have also explored planning for heterogeneous teams using both MIP techniques \cite{Koes2005,Atay2006,Lippi2021,Leahy2022} and learning-based approaches \cite{Wang2022,Paul2022}. One advantage of heterogeneous robot teams is their ability to more effectively distribute risk across the team \cite{Wu2021}.

In this paper, we present an approach that uses a graph structure and MIP to enable risk-aware planning for heterogeneous multi-robot teams under environmental uncertainty. 
% Instead of focusing on the
Contrary to the multi-robot routing problem, % \tflag{R-2-7}\blue{(which is formulated to explicitly disallow simultaneous traversing)}, 
we assume that multiple agents can simultaneously traverse an edge. Similar to \cite{Chung2019}, we model our edge costs as random variables whose values can be uncovered via exploration. \tflag{R-2-8}\blue{We employ a heterogeneous robot team, where each robot class possesses a different risk profile %to more effectively navigate in an unknown environment. The robot class's risk profile dictates its 
that dictates the class's tolerance for being detected. 
Our carrier robots are more risk-adverse, 
% due to impact of detecting the carrier vehicles being more significant than for the scout vehicles, 
since they are responsible for the underlying goal of the scenario (with the aid of scout robots) and thus their detection would be more significant. 
% Our carrier vehicles are more risk-adverse (the impact of their detection would be more significant than for the scout vehicles), since the carrier vehicles are working to accomplish the underlying goal (with the aid of the scout vehicles).
} To solve for a multi-robot plan, we use a compact MIP formulation. 
\original{In our prior research \cite{dimmig2023mip}, we used}\blindreview{In \cite{dimmig2023mip},} 
a similar approach\blindreview{ was used} to generate multi-robot plans for a homogeneous multi-robot team on a topological graph with deterministic edge costs. Here, we present an approach to address the challenges posed by coordinating a heterogeneous team in uncertain environments. 

%%%%%%%%%%%%%%%%%%%%%%%%%%%%%%%%%%%%%%%%%%%%%%%%%%%%%%%%%%%%%%%%%%%%%%%%%%%%%%%%
\section{PROBLEM STATEMENT}
\label{sec:problem_statement}
We consider the problem of planning and coordination for a heterogeneous robot team by generating a graph $G=(V, E, W_E)$, with nodes $v \in V$ and edges $e \in E$, based on the critical features of the planning problem. 
The edge costs $w_e \in W_E : E \times X \times T \rightarrow \mathbb{R}_{>0}$ are bounded random variables that depend on the underlying environmental uncertainty $u_e \in [u_e^{\llcorner}, u_e^{\ulcorner}]$, the discrete robot team state on the graph $x \in X$, and time $t\in\mathbb{R}_{>0}$. We consider $n_{CL}$ robot classes and $n_E$ total edges such that 
% $X=\{X^1,X^2, \dots, X^{n_{CL}}\}$, 
$X=X^1 \times X^2 \times \dots \times X^{n_{CL}}$,
where $X^i=\{ [p_{e_1}^i, p_{e_2}^i, \dots, p_{e_{n_E}}^i] \in \mathbb{Z}_{>0}^{n_E} \}$ and $p_{e_j}^i$ is the number of robots of class $i$ on edge $e_j$. 
We assume a level of abstraction where multiple robots may simultaneously traverse an edge. We also allow robots to uncover \blue{the true edge cost} by inspecting an edge. 
%After a robot has traversed the edge, we assume the uncertainty bounds increase with time to their original values approximately logarithmically, $\ln(k)$ for $k \in [1, \lambda]$, where $\lambda$ is the inspection decay horizon. 
%\tflag{R-2-9}\blue{The information from each inspection decays over time as the information becomes less valuable, until the 
%estimate for the 
%uncertainty %, $\hat{u}_e$, 
%returns to its original value.} 
\tflag{R-2-9}\blue{Following an inspection, we assume that the edge cost uncertainty returns to its original value after a finite time.}
%approximately logarithmically, $\ln(k)$ for $k \in [1, \lambda]$, where $\lambda$ is the inspection decay horizon. 
% \tflag{R-2-9}\blue{The uncertainty value returning to it's original value is a reflection of previous inspections becoming less valuable over time.} 

We furthermore let each robot class $i$ be constrained to graph $G^i = (V^i, E^i, W_E^i)$, where $V^i\subseteq V$, $E^i\subseteq E$, and $w_e^i \in W_E^i$ such that $w_e^i(e^i,x,t^i)=\zeta^i w_e(e^i,x,t^i)$ \tflag{R-2-10}\blue{for $e^i \in E^i$, $x \in X$, and $t^i \in [1, n_\tau^i]$}. We can now capture heterogeneous mobility, cost, speed, and sensing characteristics by defining for each robot class a graph $G^i$, an edge cost scale factor $\zeta^i$, and traversal time factor $n_\tau^i$.
%, and uncertainty bound reduction factor $\xi^i$ respectively. 
We also permit a set of costs %$J(X_t,X_{t-1},G_t)$ 
and constraints %$C(X_t,X_{t-1},G_t)$ 
that govern 
inter-class interactions.

\subsection{Reconnaissance Test Scenario}

\tflag{R-2-11}\tflag{R-1-3}\blue{
In this paper, we consider a particular reconnaissance test scenario where a heterogeneous team of robots must traverse an environment while minimizing detection by observers whose exact positions are unknown. We assume that our graph $G$ can be generated using a visibility metric computed from a distribution over observer locations, similar to \cite{dimmig2024workshop}, where nodes are assigned to low visibility regions and edge costs embed probability of detection. Thus, the distribution over observer locations gives rise to uncertain edge weights. 

%In this paper, we consider a reconnaissance scenario as a representative test case, where uncertainty comes from adversarial observers at positions unknown to the robot team. 

%We can generate our graph $G$ from continuous space by calculating visibility from the observer's expected location. Then we can assign nodes to regions of cover and edge costs based on detectability and distance, as in \cite{dimmig2024workshop}. We assume a distribution for the observer's location which results in uncertainty on the edge costs.
}
\tflag{R-2-11}\blue{
In this scenario, we assume that the observer position can change incrementally over time, such that the true edge costs uncovered from inspecting an edge have the highest certainty in the first time step following an inspection and become less certain as time progresses. Instead of modeling observer dynamics explicitly, we attempt to capture observer behavior implicitly through the temporal growth of edge cost uncertainty after an inspection. %We model the inspection information 

%We assume that our inspection information is most valuable after the first time step following an inspection (i.e., 

%We assume our inspection information is most valuable in the first time step after an inspection (i.e., the observer does not instantaneously move). We then consider this information to become less certain as time progresses and instead rely on new inspections, as we do not assume a model for the observer's movement. 
%Information from an inspection can result in an entirely new estimate for the uncertainty distribution on an edge. For this evaluation, we consider the uncertainty to remain constant. 

% Additionally, we assume the observer does not move instantaneously, such that our inspection information is most valuable in the first time step after an inspection. 
%We then consider this information to become less certain as time progresses (and instead rely on new inspections), as we do not assume a model for the observer's dynamics. 
% For this evaluation, we consider the uncertainty to be constant on an edge, however, an inspection could result in a new estimate for the uncertainty and update the value of $\hat{u}_e$ when re-planning.
}

We restrict our robot team to %consist only of 
two classes: carrier robots and scout robots. %Carrier robots are slower, incur higher cost, and are not equipped with sensors for reducing edge cost uncertainty. 
Scout robots move more quickly than carrier robots, incur less cost, and possess sensors for inspecting edges to reduce uncertainty bounds to zero. 
Carrier robots can carry at most one scout robot. Scouts can deploy when the carrier robots are at nodes and return within one carrier robot time increment to any empty carrier robot. When a scout vehicle is deployed from a carrier robot, it incurs a specified deployment cost.
In this paper, 
the edge cost scale factor for carrier robots and scouts are $1$ and $\zeta$, respectively, and the traversal time factors are $1$ and $n_\tau$, respectively.
% carrier robots operate with an edge cost scale factor and traversal time factor of $1$.
% The scout vehicle edge cost scale factor and traversal time factor are defined as $\zeta$ and $n_\tau$, respectively. 
%We also assume $E^i=E,\quad V^i=V,\quad \forall i$.
We also assume that all robots operate on graphs with identical sets of vertices and edges (i.e., $E^i=E,\quad V^i=V,\quad \forall i$). 
%Carrier robots operate with an edge cost scale factor of $1$ and we call the edge cost scale factor for scouts $\zeta$. The traversal time factors for carrier robots and scouts are $n_T$ and $n_\tau$, respectively.

% We define the relationship between the carrier and scout robots as follows. Carrier robots can carry at most one scout robot. Scouts can deploy when the carrier robots are at nodes and return within one carrier robot time increment to any empty carrier robot. When a scout vehicle is deployed from a carrier robot, it incurs a specified deployment cost.

Given the constraints of this robot team, our overall objective is to minimize the total cost incurred by the team when the edge costs are uncertain. 
% However, we must define a cost that adequately captures the underlying uncertainty. 
To do so, we define a cost that adequately captures the underlying uncertainty.
We model our uncertain edge costs using bounded uncertainty; thus, we can use the Hurwicz Criterion \cite{Loui1983, Denoeux2019} as part of our objective function. The Hurwicz Criterion was developed for decision making under interval uncertainty to balance pessimism and optimism by considering both the best and worst possible outcomes \cite{HurwiczCriterion}. We define a ``coefficient of optimism,'' $\beta \in [0,1]$, that is set based on the tolerance of the outcome. Then for a condition $c$ we want to minimize, which is bounded such that $c^{\llcorner} \leq c \leq c^{\ulcorner}$, the Hurwicz Criterion leads to the minimization of a cost of the following form.
\begin{align}
    \beta c^{\llcorner} + (1 - \beta) c^{\ulcorner} \label{eq:hurwicz}
\end{align}

For $\beta = 1$, this criterion expresses complete optimism and for $\beta = 0$, this criterion expresses complete pessimism. 

\section{MIXED-INTEGER PROGRAMMING APPROACH} \label{sec:mip}

\begin{table}[tbh]
	\centering
    \vspace*{2mm}
	\caption{MIP Parameters}
	\vspace*{-5mm}
    \label{tab:parameters}
	\begin{center}
		\renewcommand{\arraystretch}{1.3}
		\begin{tabular}{ c | c | p{4.9cm} }
			%	\hline
			%	\multicolumn{6}{|c|}{Agent} \\
			%	\hline
			%Variable Group Name & 
			\textbf{Category} & \textbf{Var} & \textbf{Description} \\
			\hline
			\hline
			\multirow{9}{4.8em}{Problem Size}
			& $n_A$ & Number of carrier agents/robots \\
            & $n_{K}$ & Number of scouts \\
			& $n_T$ & Number of time steps in the time horizon \\ 
			& $n_{\tau}$ & Number of scout time steps \\ 
			% & $n_\scriptO$ & Number of overwatch opportunities \\
			& $n_E$ & Number of edges, both directions \\
			& $n_V$ & Number of nodes/vertices \\
			& $n_L$ & Number of locations ($n_E + n_V$) \\
			& $n_S$ & Number of start locations \\
			& $n_D$ & Number of goal/destination locations \\
   			\hline
			\multirow{6}{4.8em}{Scenario Variables} 
            & $E$ & Set of edges $e$ \\
            & $V$ & Set of nodes/vertices $v$ \\
            & $L$ & Set of locations $l$ consisting of edges and vertices, $E \cup V$ \\
            & $S$ & Set of start locations $s$, $S \subseteq L$ \\
            & $D$ & Set of goal/destination locations $d$, $D \subseteq L$ \\
            % & $\scriptO$ & Set of overwatch opportunities $\scripto$, $(v_i, e_j)$ where node $v_i$ can overwatch edge $e_j$ \\
			\hline
			\multirow{4}{4.8em}{Problem Parameters} 
			& $t$ & Time step from $1$ to $n_T$ \\
            & $\tau$ & Scout time step from $1$ to $n_{\tau}$ \\
			& $n_{s}$ & Number of robots at start location $s \in S$ \\
			& $n_{d}$ & Number of robots at goal location $d \in D$ \\
			\hline
			\multirow{4}{4.8em}{Cost of Traversing} 
			& $\bar{w}_e$ & Expected cost to traverse edge $e \in E$ \\
			% & $a_e$ & Minimum desired number of robots on $e$ \\ 
			% & $m_e$ & Additional cost for robots on $e$ before $a_e$ \\
			& $r_e$ & Cost reduction on $e$ for teaming \\
            & $\zeta$ & Scout edge cost reduction \\
            & $\eta_v$ & Cost of scout launch from node $v \in V$ \\
			% \hline
			% \multirow{3}{4.8em}{Cost of Overwatch} 
			% & $\omega_\scripto$ & Benefit of full overwatch for $\scripto~\in~\scriptO$ \\
			% & $\alpha_\scripto$ & Number of robots for full overwatch for $\scripto$ \\ 
			% & $\gamma_\scripto$ & Reward for overwatch robots over $\alpha_\scripto$ for~$\scripto$ \\ 
            \hline
            \multirow{6}{4.8em}{Uncertainty} 
			& $u_e^{\llcorner}$ & Lower bound on uncertainty of edge $e \in E$ \\
            & $u_e^{\ulcorner}$ & Upper bound on uncertainty of edge $e \in E$ \\
%			& $\gamma_e$ & Reduction in uncertainty for edge $e$, $\gamma_e < u_e$ \\ 
			& $\xi$ & Scale of uncertainty for all edges versus traversed edges \\
			& $\lambda$ & Time horizon for inspections to decay \\
            & $\beta$ & Coefficient of optimism   
		\end{tabular}
	\end{center}
    \vspace*{-6mm}
    % \bigskip 
\end{table}
\begin{table}[tbh]
    \vspace*{1mm}
	\centering
	\caption{MIP Decision Variables (At Time $t$)}
    \vspace*{-4.5mm}
	\label{tab:decision_vars}
	\begin{center}
		\renewcommand{\arraystretch}{1.3}
		\begin{tabular}{ c | c | c | c | p{3.96cm} }
			%	\hline
			%	\multicolumn{6}{|c|}{Agent} \\
			%	\hline
			%Variable Group Name & 
			\textbf{Variable} & \textbf{Type} & \textbf{LB} & \textbf{UB} & \textbf{Description} \\
			\hline
			\hline
			%Location Occupancy & 
			$p_{l,t}$ & Int & 0 & $n_A$ & Number of robots at location $l$ \\ % at time $t$ \\ 
			$\phi_{e,t}$ & Bin & 0 & 1 & Whether robots are on edge $e$ \\ % at time $t$ \\
			%Time Used & 
			$\psi_{t}$ & Bin & 0 & 1 & Whether robots have moved \\ % at time $t$ \\
			%Cost of Traversing & 
			% $C_{W_{A_{e,t}}}$ & Cont. & 0 & $\infty$ & Cost of traversing edge $e$ \\ % at time $t$ \\
			%Cost of Overwatch & 
			% $C_{\Omega_{\scripto,t}}$ & Cont. & $-\infty$ & 0 & Cost of overwatch opportunity $\scripto$ \\ % at time $t$
			%	\hline
            % \hline
			% \hline
			$q_{l,\tau,t'}$ & Int & 0 & $n_{K}$ & Number of scouts at $l$ at time $\tau$ \\ % at time $t$ \\ 
            $\theta_{e',t'}$ & Bin & 0 & 1 & Whether scouts are on $e'$ at time $\tau$ \\ % at time $t$ \\ 
			$f_{v,t'}$ & Int & 0 & $n_{K}$ & Number of scouts deployed from $v$ \\ % at time $t$ \\ 
			$\delta_{e',t'}$ & Bin & 0 & 1 & Whether edge $e'$ was inspected \\ % at time $t$ \\
			$z_{e',t}$ & Cont & 0 & 1 & Inspection ratio for edge $e'$ \\ % at time $t$ \\
			% $\zeta_{e,t}$ & Cont. & 0 & 1 & Inspection ratio, robot used edge $e$\\ % at time $t$ \\
            % $\upsilon_{e,\tau,t}$ & Cont. & 0 & 1 & Inspection ratio, scout used $e$ at $\tau$ \\ % at time $t$ \\
            $\breve{C}_{U_{A_{e,t}}}$ & Cont & 0 & $\infty$ & Carrier cost of uncertainty for $e$ \\
            $\breve{C}_{U_{K_{e',t'}}}$ & Cont & 0 & $\infty$ & Scout cost of uncertainty for $e'$ 
		\end{tabular}
	\end{center}
    \vspace*{-8mm}
    % \bigskip \bigskip
\end{table}

Rather than pursuing a search-based or learning-based solution to the multi-robot planning and coordination problem, we adopt an optimization-based approach where the costs and constraints can be directly encoded. Since our problem is characterized by both discrete and continuous decision variables, we formulate the uncertainty-aware multi-robot planning problem using MIP. 
We introduce the parameters used in our formulation in Table~\ref{tab:parameters}. A scenario is defined by the categories ``Problem Size,'' ``Scenario Variables,'' and ``Problem Parameters'' based on the environment of interest. These categories encompass the parameters relating to the graph connectivity, start and goal locations of the robots, and planning horizons. The ``Cost of Traversing'' and ``Uncertainty'' categories define parameters that will affect the team's behaviors on the graph %(e.g., edge cost, edge uncertainty, teaming benefits, uncertainty decay) 
and are based on the observer's expected location. In this work, we select values of these parameters to demonstrate the behaviors of our algorithm. 

%In this formulation, we add a new type of robot, scouts, which can move faster than the carrier robots to collect information to minimize uncertainty about graph edges. 
Our total number of robots is $n_A + n_K$, carriers and scouts. To represent the accelerated speed of the scouts, the scouts operate with time step $\tau$. During one time step $t$ for the carrier robots there are $n_{\tau}$ scout time steps $\tau$. %, allowing the scouts to travel further than the carrier robots. 

Table~\ref{tab:decision_vars} presents the categories of decision variables we employ as well as their type (integer, binary, or continuous), bounds (lower (LB) and upper (UB)), and descriptions. 
The key decision variables $p_{l,t}$ for carrier robots and $q_{l,\tau, t}$ for scout robots %describe the positions of each class of robots %: $p_{l,t}$ for carrier robots and $q_{l,\tau, t}$ for scout robots. 
track the number of robots at a particular location $l$ at each time step $t$ (and for scouts, scout time step $\tau$). After a solution is generated to the MIP problem, an assignment routine can be performed to construct the path each robot takes from these variables.

For an undirected graph, we consider both directions of each edge in $E$ to be equivalent. %, with equivalent expected cost and uncertainty. % of each edge are assumed to be the equivalent for both directions of the edge. % and thus inspecting an edge in one direction is equivalent to inspecting the edge in the other direction. 
In some cases, we can consider the set of only one direction of each edge, $E'$, with element $e' \in E'$.
To remove extraneous variables, scout decision variables exclude the last time step $t = n_T$ and potentially the first time step if their deployment is delayed. We use $t'$ to distinguish cases with a truncated time horizon.  
% Overall these factors reduce the number of required variables. 
In Table~\ref{tab:decision_vars}, we denote these cases with $e'$ and $t'$. %, however,  
For brevity, we drop this notation in our derivation. 

% To optimize solve time we formulate out cost functions and constraints to be linear, ultimately yielding a Mixed Integer Linear Program (MILP).
% To preserve linearity, when a nonlinear expression is desired in the formulation we introduce linear inequality constraints that set decision variables to the desired values by using their role in the cost function minimization. Ultimately these values end up equating to the originally desired nonlinear expression using constraints entirely linear in the decision variables. An example of this is our formulation of constraints for the carrier and scout costs of uncertainty $C_{U_{A_{e,t}}}$ and $C_{U_{K_{e,t}}}$ in Section~\ref{sec:cost_trav_uncertainty}. 
% Additionally, 
% we introduce binary decision variables, as seen in Table~\ref{tab:decision_vars} for linearity of expressions. These binary decision variables are entirely dependent on other decision variables. For example, $\phi_{e,t} = 0$ if $p_{e,t} = 0$ and $\phi_{e,t} = 1$ if $p_{e,t} > 0$. The purpose of these decision variables will be described further in the following sections.

\subsection{Mathematical Preliminaries}
\label{sec:math_prelim}

In this work, we add slack variables and constraints to express nonlinear terms of our cost functions linearly, ultimately yielding a Mixed-Integer Linear Program (MILP). We strive to minimize the number of additional variables and constraints added since these impact the overall solve time of the MILP.
To do this, we exploit the conclusions in \cite{Glover1975, Adams2005} to transform a quadratic cost term into a linear cost term with linear constraints using a slack variable. 
\begin{proposition}
If $\alpha \in \{0,1\}$ is a binary variable, $b \in \mathbb{R}$, and $h(b)$ is a linear function with bounds $h^{\llcorner} \leq h(b) \leq h^{\ulcorner}$, then minimizing $\alpha h(b)$ is equivalent to $\min y$ for slack variable $y \in \mathbb{R}$ with the following constraints.
\begin{align}
    \alpha h^{\llcorner} \leq &y \leq \alpha h^{\ulcorner} \\
    h(b) - h^{\ulcorner} (1 - \alpha) \leq &y \leq h(b) - h^{\llcorner} (1 - \alpha)
\end{align}
\label{math:prop:quad_to_linear}
\end{proposition}
\vspace*{-8mm}
\begin{corollary}
When minimizing, if all other constraints are independent of $y$ and the coefficient of $y$ in the cost function is nonnegative, then only the left side constraints are structural, and thus necessary for optimization. 
\begin{align}
    y &= \max\{\alpha h^{\llcorner}, h(b) - h^{\ulcorner} (1 - \alpha)\} \\
    &\leq \min\{\alpha h^{\ulcorner}, h(b) - h^{\llcorner} (1 - \alpha) \}
\end{align}
\label{math:coro:quad_to_linear}
\end{corollary}
\vspace*{-8mm}
\begin{specialcase}    
For a constant $a \geq 0$, $h(b) = 1 - b$, $b \in [0,1]$, and thus $h^{\llcorner} = 0$ and $h^{\ulcorner} = 1$, the minimization of $a \alpha (1 - b)$ can be equivalently expressed as follows.
\begin{align}
    \min~ y ~
    \text{s.t.}~ y \geq a (\alpha - b),~ y \geq 0
\end{align}
\label{math:sc:quad_to_linear}
\end{specialcase}
\vspace*{-8mm}
\begin{proposition} 
For $b \in [0, b^{\ulcorner}]$, an indicator variable $\alpha \in \{0,1\}$ can be expressed as follows, for any constant $a \geq 0$. 
\begin{align}
    \alpha = \begin{cases}
        1,~~ b > 0 \\
        0,~~ b = 0
    \end{cases}
    ~~~
    \Leftrightarrow
    ~~~~~
    \min a \alpha~\text{s.t.}~\alpha \geq \frac{b}{b^{\ulcorner}}
\end{align}
\label{math:prop:indicators}
\end{proposition}
\vspace*{-6mm}

%%%%%%%%%%%%%%%%%%%%%%%%%%%%%%%%%%%%%%%%%%
\subsection{Heterogeneous Team and Uncertainty Cost Functions}
%%%%%%%%%%%%%%%%%%%%%%%%%%%%%%%%%%%%%%%%%%

The main objective in our reconnaissance test scenario is to minimize detection by observers and the time to reach a goal location(s). 
We represent a vehicle's detectability through edge costs incurred while traversing the graph, with associated uncertainty values to represent the stochasticity in our prediction of the observer's location. %Scouts can be launched to gather information and reduce that uncertainty. 
% To achieve our objective 
We compose a cost function with four categories, which we derive in the following sections: traversing ($C_{W_{e,t}}$), uncertainty ($C_{U_{e,t}}$), launching scouts ($C_{F_{v,t}}$), and time ($C_{T_t}$). We sum these terms across all time steps to represent the overall objective function value $C$, which we aim to minimize. Each of these terms can be scaled based on the priorities of a scenario (e.g., minimizing detection versus the time to reach the goal). 
% The following sections outline the formulation of each cost. 
\begin{align}
    C = \sum_{t = 1}^{n_T} \bigg( C_{T_t} + &\sum_{e \in E} \big( C_{W_{e,t}} + C_{U_{e,t}} \big) + \sum\limits_{v \in V} C_{F_{v,t}} \bigg) \label{eq:overall_cost}
\end{align}

\subsubsection{Cost of Traversing and Uncertainty}
\label{sec:cost_trav_uncertainty}
% In \cite{dimmig2023mip}, we formulated a piecewise-linear cost of traversing based on an expected base cost to traverse an edge, $\bar{w}_e$, 
%the desired number of robots on the edge, and 
% with incentives for teaming. 
%We add additional cost to incentivize reaching the desired number of robots and then cost reductions for additional robots as a further incentive for teaming. 
We first consider, for each time step $t$, the uncertain cost of traversing edge $e$, $\tilde{C}_{W_{e,t}}$, that is equal to the weight on the edge, $w_e$, if any carrier vehicles are on that edge plus a scaled weight $\zeta w_e$ for each scout on that edge. We consider $w_e$ to be a random variable with an expected value of $\bar{w}_e$ and interval uncertainty. We add the scaling factor $\zeta$ to represent a difference in cost for scouts versus carrier vehicles to traverse the edge (e.g., scouts movement may be less risky due to their increased speed). We apply the cost $w_e$ once for carrier robots since they traverse as a team and, when traversing separately, $\zeta w_e$ for each scout vehicle because they move independently.  
We track whether carrier vehicles are traversing edge $e$ with the binary variable $\phi_{e,t}$ and, for each scout time step $\tau$, how many scouts are traversing edge $e$ with integer variable $q_{e,\tau,t}$. 
%Additionally, we scale $w_e$ by a parameter $\zeta$ to represent a potential difference in cost for scouts to traverse the edge compared to the carrier vehicles. 
%
% the which is a linear cost that depends on the weight on the edge, $w_e$. This cost applies to the carrier robots if they use that edge
% % , and whether that edge is used by the carrier robots 
% at time $t$ (i.e., $\phi_{e,t}$) and we add a term for each scout that uses that edge at scout time step $\tau$ and overall time step $t$ (i.e., $q_{e,\tau,t}$) with the cost reduced by $\zeta$ to represent the lower cost for scout's to traverse. 
% . Additionally, we add a term for the cost of traversing for scouts using the number of scouts on an edge $e$ at scout time step $\tau$ and overall time step $t$, i.e. $q_{e,\tau,t}$. 
%
\begin{align}
    \tilde{C}_{W_{e,t}} %&= w_e \phi_{e,t} + \sum_{\tau = 1}^{n_{\tau}} w_e q_{e,\tau,t} \\
    &= w_e \bigg(\phi_{e,t} + \zeta \sum_{\tau = 1}^{n_{\tau}} q_{e,\tau,t} \bigg)
    \label{eq:simplified_cw}
\end{align}

%We consider $w_e$ to be a random variable with an expected value of $\bar{w}_e$ and interval uncertainty. %associated uncertainty $u_e$ (e.g., a 95\% confidence interval on a Gaussian distribution). 
Let the upper bound of $w_e$ be $\bar{w}_e + u_e^{\ulcorner}$ and the lower bound be $\bar{w}_e - u_e^{\llcorner}$. 
We can then apply (\ref{eq:hurwicz}) to define $\hat{w}_e = \beta (\bar{w}_e - u_e^{\llcorner}) + (1 - \beta) (\bar{w}_e + u_e^{\ulcorner})$.
% We then apply %the Hurwicz Criterion from <-- incorrect language
% (\ref{eq:hurwicz}) to represent $w_e$.
%
% \begin{align}
%     \hat{w}_e &= 
%     \beta (\bar{w}_e - u_e^{\llcorner}) + (1 - \beta) (\bar{w}_e + u_e^{\ulcorner}) \label{eq:uncertain_weights}
%     % &= \bar{w}_e + u_e(1 - 2\beta)
%     % &= \bar{w}_e + (1 - \beta) u_e^{\ulcorner} - \beta u_e^{\llcorner}
% \end{align}

After reorganizing terms, this results in $\bar{w}_e$ plus an expression dependent on the uncertainty bounds, $\hat{u}_e = (1 - \beta) u_e^{\ulcorner} - \beta u_e^{\llcorner}$. By letting $w_e \approx \hat{w}_e$, we can derive a cost based on (\ref{eq:simplified_cw}) that, if minimized, satisfies the Hurwicz Criterion. 
% For simplicity, we call this expression for the uncertainty $u_e = (1 - \beta) u_e^{\ulcorner} - \beta u_e^{\llcorner}$.
%
\begin{align}
    \hat{\tilde{C}}_{W_{e,t}} %&= (\bar{w}_e + u_e) \bigg(\phi_{e,t} + \sum_{\tau = 1}^{n_{\tau}} q_{e,\tau,t} \bigg) \label{eq:C_trav_tilde} \\
    % \begin{split}
    &= \bar{w}_e \bigg(\phi_{e,t} + \zeta \sum_{\tau = 1}^{n_{\tau}} q_{e,\tau,t} \bigg) %\\ %\label{eq:c_trav_bar}\\ 
    % & \quad \quad \quad 
    + \hat{u}_e \bigg(\phi_{e,t} + \zeta \sum_{\tau = 1}^{n_{\tau}} q_{e,\tau,t} \bigg) %\label{eq:c_trav_u}
    % \end{split}
    \label{eq:C_trav_tilde2}
\end{align}

\textbf{Cost of Traversing:}
The first term in (\ref{eq:C_trav_tilde2}) is an expression of the expected cost of traversing. % for the simplified cost expression, $\tilde{C}_{W_{e,t}}$. %, and we label this term $\bar{\tilde{C}}_{W_{e,t}}$. 
In our final cost of traversing $C_{W_{e,t}}$, we add a linear cost reduction $r_e$ based on the number of agents on the edge $p_{e,t}$ as a risk reduction when multiple carrier vehicles traverse together, as in \cite{dimmig2023mip}. 
\begin{align}
C_{W_{e,t}} &= \bar{w}_e \phi_{e,t} - r_e p_{e,t} + \zeta \bar{w}_e \sum_{\tau = 1}^{n_{\tau}} q_{e,\tau,t} \label{eq:total_cost_traversing} %\\
% C_{W_{A_{e,t}}} &\geq - m_e p_{e, t} + (\bar{w}_e + m_e a_e) \phi_{e,t} \label{eq:cost:trav1} \\
% C_{W_{A_{e,t}}} &\geq - r_e p_{e, t} + (\bar{w}_e + r_e a_e) \phi_{e,t} \label{eq:cost:trav2} 
\end{align}

\textbf{Cost of Uncertainty:}
The second term in (\ref{eq:C_trav_tilde2}) is the impact of the uncertainty $u_e$ on the cost of traversing edge $e$ at time $t$. 
% we label this term $C_{U_{W_{e,t}}}$. 
% In addition to this cost of uncertainty, %to the uncertainty on the edges being traversed, 
We add a term to this expression for the total uncertainty across all edges, scaled by $\xi$, to encourage general exploration by the scouts (rather than focused exploration on edges intended to be traversed), in case new information would reveal a superior path. % through the environment. 
% We scale the additional uncertainty term 
% by $\xi$ to reflect the priority of general exploration versus exploration on the edges that are planned to be used. 
Together, these terms form an expression for the maximum cost of uncertainty as follows. 
% We add to this combine this term with the uncertainty on all edges to express the total cost of uncertainty. 
% To express the total cost of uncertainty we combine $C_{U_{W_{e,t}}}$ with the uncertainty on all edges. 
% We then formulate our total cost of uncertainty from the uncertainty of traversed edges, $C_{U_{W_{e,t}}}$, and the uncertainty on all edges. 
% We consider the uncertainty across the graph to encourage general exploration, and scale it by $\xi$ to reflect the priority of general exploration versus exploration on the edges that are planned to be used.
% This expression is as follows and represents the maximum cost of uncertainty.
%
\begin{align}
    \hat{u}_e \bigg(\xi + \phi_{e,t} + \zeta \sum_{\tau = 1}^{n_{\tau}} q_{e,\tau,t} \bigg) \label{eq:uncertainty_expression}
\end{align}

% The expression in (\ref{eq:uncertainty_expression}) represents the maximum cost of uncertainty. 
The role of the scouts in our formulation is to reduce uncertainty by exploring edges ahead of the rest of the team. 
To reflect this, we add a ratio $z_{e,t}$ for how recently an edge has been inspected. We use this ratio to reduce the uncertainty of explored edges (i.e., the incurred cost of uncertainty is $\hat{u}_e (1 - z_{e,t})$). The ratio $z_{e,t}$ is $0$ if the edge has not yet been inspected and $1$ when the edge was inspected at $t-1$. The value of $z_{e,t}$ decays over time to reflect how recently the scouts explored an edge. %an edge has been explored by the scouts. 
Additionally, when running with a receding horizon, the collected data can be used to update the values of the edge weights and uncertainty. % can be updated based on the scout's data. %new data collected by the scouts. 

During a carrier robot time step $t$, after scouts explore an edge for the first time, %during their intermediate time steps, 
the scouts no longer incur the cost of uncertainty on additional passes of that particular edge during time $t$. The scouts gathered full knowledge of that edge for time $t$ in their first pass. %first pass gave full knowledge of the edge for time $t$. 
To enforce this paradigm, we replace $\sum_{\tau = 1}^{n_{\tau}} q_{e,\tau,t}$ in (\ref{eq:uncertainty_expression}) with decision variable $\theta_{e,t}$, which tracks whether scouts use an edge $e$ at time $t$.
% Thus we only consider uncertainty on the scouts first pass of an edge by tracking whether scouts use an edge $e$ at time $t$ with decision variables $\theta_{e,t}$.
% Since the scouts are reducing the uncertainty to $0$ after exploration, we only consider the uncertainty for the scouts on their first pass of an edge. We introduce decision variables $\theta_{e,t}$ that track whether scouts used edge $e$ at time $t$. 
%These ratios will be explained in more detail in Section XXX. These ratios, $z_{e,t}$, $\zeta_{e,t}$, $\upsilon_{e,\tau, t}$, are to reduce the uncertainty across all explored edges, across used edges, and across scout used edges, respectively. 

We combine the reduction in uncertainty due to $z_{e,t}$ and the use of $\theta_{e,t}$ with the expression in (\ref{eq:uncertainty_expression}) to calculate our overall cost of uncertainty for a particular edge $e$ at time $t$.
\begin{align}
    C_{U_{e,t}} = \hat{u}_e (\xi(1 - z_{e,t}) + (\phi_{e,t} + \zeta \theta_{e,t}) (1 - z_{e,t}) ) %+ \theta_{e,t} (1 - z_{e,t}) 
    \label{eq:cost_uncertainty}
\end{align}

% \begin{align}
%     \begin{split}
%     C_{U_{e,t}} = u_e \bigg(\xi(1 - z_{e,t}) + \phi_{e,t} - \zeta_{e,t} \\
%     \quad \quad + \sum_{\tau = 1}^{n_{\tau}} (\theta_{e,\tau,t} - \upsilon_{e,\tau, t})\bigg)
%     \end{split}
% \end{align}

% Note: $\zeta_{e,t} = z_{e,t} \phi_{e,t}$. I.e. $z_{e,t} = \zeta_{e,t}$ for an edge that is used. If the edge is not used $\zeta_{e,t} = 0$. The variable $\zeta_{e,t}$ is created to keep the cost function linear. Linear constraints are added to enforce the relation to $z_{e,t}$ and $\phi_{e,t}$.

The expression in (\ref{eq:cost_uncertainty}) is nonlinear in the decision variables. To formulate (\ref{eq:cost_uncertainty}) as a linear cost with linear constraints, we first break this expression into a linear component and two cost terms $C_{U_{A_{e,t}}}$ and $C_{U_{K_{e,t}}}$. %cost constraints as follows.
\begin{align}
    C_{U_{e,t}} &= \hat{u}_e \xi(1 - z_{e,t}) + C_{U_{A_{e,t}}} + C_{U_{K_{e,t}}} \label{eq:total_cost_uncertainty} \\
    C_{U_{A_{e,t}}} &= \hat{u}_e \phi_{e,t}(1 - z_{e,t}) \label{eq:robot_cost_uncertainty_eq}\\
    C_{U_{K_{e,t}}} &= \zeta \hat{u}_e \theta_{e,t}(1 - z_{e,t}) \label{eq:scout_cost_uncertainty_eq}
    % C_{U_{A_{e,t}}} &\geq u_e (\phi_{e,t} - z_{e,t}) \label{eq:robot_cost_uncertainty}\\
    % C_{U_{K_{e,t}}} &\geq \zeta u_e (\theta_{e,t} - z_{e,t}) \label{eq:scout_cost_uncertainty}
\end{align}

% Constraints (\ref{eq:robot_cost_uncertainty}) and (\ref{eq:scout_cost_uncertainty})
% We first assume $u_e \geq 0$ for all edges $e$. 
% Our calculation of the uncertainty, as described in Section XXX, ensures that $u_e^{\llcorner} < u_e^{\ulcorner}$, so this assumption only requires that we set $\beta < 0$, which is desirable in our test case since we aim to be risk averse. 
In our test case, our aim is to be risk-averse, so we can assume $u_e^{\llcorner} < u_e^{\ulcorner}$ and $\beta < 0.5$ to ensure $\hat{u}_e \geq 0$ for all edges~$e$. 
% In other scenarios, we would add additional constraints to remove these assumptions. 
We can then apply Special~Case~\ref{math:sc:quad_to_linear}, by using the cost terms $\breve{C}_{U_{A_{e,t}}}$ and $\breve{C}_{U_{K_{e,t}}}$ as slack variables with lower bounds of $0$. We add the following constraints.
% Next, we add the cost terms $C_{U_{A_{e,t}}}$ and $C_{U_{K_{e,t}}}$ as decision variables with lower bounds of $0$. 
% We can then use the following two linear constraints in place of the nonlinear equations in (\ref{eq:robot_cost_uncertainty_eq}) and (\ref{eq:scout_cost_uncertainty_eq}).
\begin{align}
    % C_{U_{e,t}} &= \hat{u}_e \xi(1 - z_{e,t}) + \breve{C}_{U_{A_{e,t}}} + \breve{C}_{U_{K_{e,t}}} \label{eq:total_cost_uncertainty_constr} \\
    \breve{C}_{U_{A_{e,t}}} &\geq \hat{u}_e (\phi_{e,t} - z_{e,t}) \label{eq:robot_cost_uncertainty}\\
    \breve{C}_{U_{K_{e,t}}} &\geq \zeta \hat{u}_e (\theta_{e,t} - z_{e,t}) \label{eq:scout_cost_uncertainty}
\end{align}
Following Special~Case~\ref{math:sc:quad_to_linear}, %since we are minimizing cost in our optimization problem, 
$\breve{C}_{U_{A_{e,t}}}$ and $\breve{C}_{U_{K_{e,t}}}$ will be strict to the minimum possible values given the constraints, and thus can replace (\ref{eq:robot_cost_uncertainty_eq}) and (\ref{eq:scout_cost_uncertainty_eq}).  
% Using constraint (\ref{eq:robot_cost_uncertainty}) as an example,
% when an edge is not traversed by the carrier robots (i.e., $\phi_{e,t} = 0$) the cost of uncertainty from the carrier robots using that edge, $C_{U_{A_{e,t}}}$, should be zero. In this case, the constraint in (\ref{eq:robot_cost_uncertainty}) 
% is $C_{U_{A_{e,t}}} \geq - u_e z_{e,t}$. Since $u_e \geq 0$ and $z_{e,t} \geq 0$, this constraint is a nonpositive bound. 
% However, we additionally have the constraint that $C_{U_{A_{e,t}}} \geq 0$. Thus for both constraints to be true, and choosing the minimum possible value since we are minimizing cost, $C_{U_{A_{e,t}}} = 0$ which is equivalent to (\ref{eq:robot_cost_uncertainty_eq}) for $\phi_{e,t} = 0$.
% % Since the lower bound on 
% % would be a negative bound, but the overall bound on $C_{U_{A_{e,t}}} \geq 0$, as stated in Table~\ref{tab:decision_vars}, would set the value of $C_{U_{A_{e,t}}}$ to $0$ since we're minimizing cost. 
% When the edge is traversed (i.e., $\phi_{e,t} = 1$), (\ref{eq:robot_cost_uncertainty}) is a nonnegative bound (i.e., $ C_{U_{A_{e,t}}} \geq u_e (1 - z_{e,t})$) and thus $u_e (1 - z_{e,t})$ is the value of $C_{U_{A_{e,t}}}$ since we are minimizing cost, which is equivalent to (\ref{eq:robot_cost_uncertainty_eq}) for $\phi_{e,t} = 1$. 
% These bounding properties are the same for $C_{U_{K_{e,t}}}$ in (\ref{eq:scout_cost_uncertainty}).
Overall, this constraint formulation allows the cost of uncertainty to remain linear with linear constraints, which significantly improves solve time. 

% \subsubsection{Cost of Traversing}

% % We can then revisit the cost of traversing. 
% From our expression for the risk-aware cost of traversing in (\ref{eq:C_trav_tilde2}), the uncertainty component is captured with $C_{U_{e,t}}$ so we can separately express the component for the expected cost of traversing. % from (\ref{eq:C_trav_tilde2}). 
% % We use the formulation from \cite{dimmig2023mip} that captures the teaming considerations in () and () and expresses the cost as a linear term in the cost function with two linear constraints and add the cost of the scouts traversing. 
% % We use the cost constraints (\ref{eq:cost:trav1}) and (\ref{eq:cost:trav1}) to capture the teaming constraints, as derived in \cite{dimmig2023mip}, and add the linear cost of the scouts traversing as follows.
% Additionally, we add a linear cost reduction based on the number of agents on an edge to incentivize teaming, as in \cite{dimmig2023mip}. 
% %
% \begin{align}
% C_{W_{e,t}} &= \bar{w}_e \phi_{e,t} - r_e p_{e,t} + \zeta \bar{w}_e \sum_{\tau = 1}^{n_{\tau}} q_{e,\tau,t} \label{eq:total_cost_traversing} %\\
% % C_{W_{A_{e,t}}} &\geq - m_e p_{e, t} + (\bar{w}_e + m_e a_e) \phi_{e,t} \label{eq:cost:trav1} \\
% % C_{W_{A_{e,t}}} &\geq - r_e p_{e, t} + (\bar{w}_e + r_e a_e) \phi_{e,t} \label{eq:cost:trav2} 
% \end{align}

\subsubsection{Cost of Launch}

To represent the risk associated with deploying a scout vehicle (e.g., an aerial vehicle may be noticeable and noisy), we introduce a cost for launching the vehicle, $C_{F_{v,t}}$. 
The launch cost for a particular node, $\eta_v$, is the average cost of the surrounding edges multiplied by a scaling factor. 
%This is only considered for linked vehicles. Linked vehicles are when the scouts must launch and return to eligible UGVs. In those cases we include this "cost of launch" for deploying from the linked vehicle. When the vehicles are not linked, the scouts can go anywhere and do not need to return. 
We express this cost for all $t$ and $v$.
\begin{align}
C_{F_{v,t}} = \eta_v f_{v,t} \label{eq:launch_cost}
\end{align}

% \subsection{Cost of Overwatch}

% Restate? 

\subsubsection{Cost of Time}

To quantify the risk associated with the time to reach a goal, we formulate a linear cost of time $C_{T_t}$ for each time step $t$, as in \cite{dimmig2023mip}. We introduce binary decision variables $\psi_t$ that track whether any robots have moved on the graph at time $t$ and scale this value by the current time $t$, such that the cost increases the longer the robots are traversing. 
% We use the cost of time as derived in \cite{dimmig2023mip} using decision variables $\psi_t$ to track movement on the graph.
%
\begin{align}
C_{T_t} = t \psi_t \label{eq:cost_time}
\end{align}

% \subsubsection{Total Cost}

% We combine our cost expressions from (\ref{eq:total_cost_uncertainty}), (\ref{eq:total_cost_traversing}), (\ref{eq:launch_cost}), and (\ref{eq:cost_time}) into our overall objective function. 
% % Constraints associated with the cost of overwatch, $C_{\Omega_{\scripto,t}}$, are reproduced for completeness in Table~\ref{tab:optimization_problem}.
% %
% \begin{align}
% % \begin{split}
% C = C_T + \sum_{t = 1}^{n_T} \bigg( &\sum_{e \in E} \big( C_{W_{e,t}} + C_{U_{e,t}} \big) + \sum\limits_{v \in V} C_{F_{v,t}} \bigg) \label{eq:overall_cost}
% % \end{split}
% \end{align}

% Each of these terms can be scaled based on the priorities of a scenario. 

%%%%%%%%%%%%%%%%%%%%%%%%%%%%%%%%%%%%%%%%%%
\subsection{Scout and Uncertainty Constraints}
%%%%%%%%%%%%%%%%%%%%%%%%%%%%%%%%%%%%%%%%%%

% We expand on the constraints from \cite{dimmig2023mip} for the new class of robots (scouts) and inclusion of uncertainty. 
% We add the following constraints when expanding our MIP formulation from \cite{dimmig2023mip} for scouts and incorporating uncertainty.

\subsubsection{Time Tracking Variables}
We define a constraint for the time tracking variables $\psi_t$. % to track whether carrier robots and scouts are moving on edges for each time step $t$. 
% At time $t$, we sum the number of carrier robots on edges, $p_{e,t}$, and the number of scouts on edges, $q_{e,\tau,t}$. 
% All edges have weight associated with them, so 
A robot would not pause on an edge due to the edge's weight, thus being on an edge implies movement. 
% We divide by the maximum value of the sum such that the right hand side (RHS) expression in the following inequality is $1$ in the case of maximum movement and 0 when agents did not move. 
% Since $\psi_t$ is binary the inequality ensures that any number of moving agents sets $\psi_t = 1$. Furthermore, since $\psi_t$ contributes to increases costs, when the RHS is $0$, $\psi_t$ will be $0$.
In the overall cost function, $\psi_t$ only contributes to increasing cost, so we constrain $\psi_t$ using Proposition~\ref{math:prop:indicators}, such that $\psi_t = 1$ when any number of robots move on an edge and otherwise $\psi_t = 0$.
\begin{align}
\psi_t \geq \frac{1}{n_A + n_K n_{\tau}} \sum_{e \in E} \bigg( p_{e,t} + \sum_{\tau = 1}^{n_{\tau}} q_{e,\tau,t} \bigg) \label{eq:constr:time_vars}
\end{align}

\subsubsection{Carrier Robot and Scout Edge Used Variables}

We define binary variables $\phi_{e,t}$ and $\theta_{e,t}$ to track if edge $e$ is being traversed by the carrier robots and scouts, respectively, at time $t$. Since $\phi_{e,t}$ and $\theta_{e,t}$ contribute to increasing the cost function, we define constraints using Proposition~\ref{math:prop:indicators}. % from Sec.~\ref{sec:math_prelim}. 
% We set the binary variable $\theta_{e,t}$ to 
% $1$ if there are scouts on edge $e$ during any scout time $\tau$ and overall time $t$ and $0$ otherwise. 
% These constraints assume $\phi_{e,t}$ and $\theta_{e,t}$ contribute to increasing the cost function, such that they will only be set to $1$ when 
% required by the inequality and will otherwise be $0$. 
% 
\begin{align}
    \phi_{e,t} \geq \frac{1}{n_A} p_{e,t}, \quad %\label{eq:carrier_edge_used} \quad
    \theta_{e,t} \geq \frac{1}{n_K n_{\tau}} \sum_{\tau = 1}^{n_{\tau}} q_{e,\tau,t} \label{eq:edge_used} %\label{eq:scout_edge_used}
\end{align}

\subsubsection{Start and Destination Constraints}

We add constraints for the start and destination locations of the overall robot team using the carrier robot locations. %, which is controlled by the carrier robot locations. 
The number of robots at each start location $s$ is $n_s$ and the minimum number of robots at each destination location $d$ is $n_d$. %This yields the following constraints.
\begin{align}
    p_{s, 1} = n_{s}, \quad %\label{eq:start_constr}
    p_{d, n_T} \geq n_{d} \label{eq:start_goal_constr} %\label{eq:goal_constr}
\end{align}

\subsubsection{Scout Deployment Constraints}

% We consider the scouts to be linked to carrier robots such that 
Scouts must launch from a carrier robot and return to a carrier robot without a scout.
%(which may be a different vehicle than that scout launched from). 
We use decision variables $f_{v, t}$ to track the number of scouts deployed from a particular location $v$ at time step $t$. We bound this value based on the number of carrier robots at that location. Additionally, we add start and goal constraints to deploy and return to those locations. For all $t$ and $v$, we add the following constraints.
\begin{align}
f_{v, t} \leq p_{v, t}, \quad q_{v, 1, t} = f_{v,t}, \quad q_{v, n_\tau, t} = f_{v, t}\label{eq:scout_deploy}
\end{align}

% \subsubsection{Scout Start Constraints}

% % We constrain the start locations of the scouts to locations a vehicle has been deployed from. 
% For all $t$ and $v$, we constrain the scouts to start from a deploy location.
% \begin{align}
% q_{v, 1, t} = f_{v,t} \label{eq:scout_start_deploy}
% \end{align}

% % For planning with a receding horizon, we set the scouts to not launch in the first time step $t=1$ such that the decision variables at $t=1$ replicate the current state of the team. Thus, for all start locations $s$ and $\tau$, we bound the starting locations of the scouts. 
% % % $\tau \in [2,n_\tau]$

% % \begin{align}
% % q_{s, \tau, 1} = n_s \label{eq:scout_start_t1}
% % \end{align}

% \subsubsection{Scout Goal Constraints}

% We ensure the scouts return to a location a scout had been deployed from, such that each carrier robot starts and ends with a scout. For each $t$ and $v$, we constrain the scout final locations. 
% %
% \begin{align}
% q_{v, n_\tau, t} = f_{v, t} \label{eq:scout_goal}
% \end{align}

\subsubsection{Maximum Robots}

We constrain the maximum number of carrier robots across all locations at all time $t$ to be equal to the total number of carrier robots. 
Similarly, we constrain the maximum number of scouts across all locations at all $\tau$ and $t$ to be equal to the number of scouts deployed.
\begin{align}
\sum\limits_{l \in L} p_{l,t} = n_A, \quad %\label{eq:max_carrier} \\
\sum_{l \in L} q_{l, \tau, t} = \sum_{v \in V} f_{v, t} \label{eq:max_robots} %\label{eq:max_scouts}
\end{align}

\subsubsection{Carrier Robot and Scout Sequential Flow Constraints}

To restrict the carrier robots and scouts movement to the topological graph, we add flow constraints for each vehicle type. % we add a constraint for movement between nodes to be along the edges between them. 
For each node $v_j$, we constrain the number of robots entering the node and in the node to be equal to the number of robots in the node and leaving the node in the next time step. 
% This ensures the robots' movements flow sequentially through the graph. 
For the carrier robots, we add constraint (\ref{eq:carrier_sequential}) for all $t \in [2, n_T]$ and node $v_j$. For the scouts, we add constraint (\ref{eq:scout_sequential}) for all $t$, $\tau \in [2, n_\tau]$, and node~$v_j$. 
\begin{align}
\sum_{\substack{l_{ij} = (v_i, v_j) \in L}} p_{l_{ij},t-1} &= \sum_{\substack{l_{ji} = (v_j, v_i) \in L}} p_{l_{ji},t} \label{eq:carrier_sequential} \\
\sum_{\substack{l_{ij} = (v_i, v_j) \in L}} q_{l_{ij},\tau-1, t} &= \sum_{\substack{l_{ji} = (v_j, v_i) \in L}} q_{l_{ji},\tau, t}  \label{eq:scout_sequential}
\end{align}

In each constraint, the first sum tracks the number of robots on edges entering node $v_j$ (i.e., locations of the form $l_{ij} = (v_i, v_j)$) and the second sum tracks the number of robots on edges exiting $v_j$ (i.e., locations of the form $l_{ji} = (v_j, v_i)$). Both sets of locations include the node (i.e., location $l_{jj}$), as all nodes have self-loops. Since the cost on the edges is greater than zero, the cost function ensures the robots will not stay on an edge, and thus being on an edge implies that the robot is headed to the corresponding node. 
%Ultimately, this property allows constraints (\ref{eq:carrier_sequential}) and (\ref{eq:scout_sequential}) to enforce movement following the edges of the graph. 

\subsubsection{Edge Inspection Variables}

We add constraints for the binary decision variables $\delta_{e,t}$ to track which edges the scouts have traversed to inspect. %have been inspected due to the traversal of a scout. 
For $e$ and $t = [1, n_T - 2]$, $\delta_{e,t}$ tracks if edge $e$ has been inspected at time $t$.
\begin{align}
\delta_{e,t} \leq \sum_{\substack{\tau = 1}}^{n_\tau} {q_{e, \tau, t}} \label{eq:edge_inspected}
\end{align}

We use $z_{e,t}$ to track inspection ratios. If edge $e$ was inspected in the last time step then $z_{e,t} = 1$. Over time $z_{e,t}$ will decay, reflecting uncertainty increasing since the last time edge $e$ was inspected. We denote $\lambda$ as the number of time steps for the inspection information to decay.

We consider the uncertainty to increase logarithmically with time. By using the additive property of logarithms over our inspection horizon, our formulation ensures there is value from each inspection to encourage more frequent exploration. 
We approximate $\ln (k)$ with the harmonic number $H_k = \sum_{n=1}^{k} \frac{1}{n}$.
%
% If edge $e$ was only inspected in the current time step $t$ then $z_{e,t} = 0$, but $z_{e,t+1}$ will be $1$.
%For receding horizon, these values are initialized based on the last step. 
Using the series identity for the harmonic number, we can isolate the impact of an inspection at each time step in the inspection horizon. %, which we use as our inspection fraction, $z_{e,t}$. 
\begin{align}
    \frac{1}{\lambda} \sum_{k=1}^\lambda H_k &= \frac{1}{\lambda} ((\lambda + 1) H_\lambda - \lambda) 
    % &= (1 + \frac{1}{\lambda}) H_\lambda - 1 \\
    = \sum_{t_d=1}^\lambda \frac{\lambda - t_d + 1}{\lambda t_d} \label{eq:lambda}
\end{align}

We then switch coordinates of (\ref{eq:lambda}) to be relative to time $t$ and use the binary $\delta_{e,t}$ to sum components corresponding to completed inspections. 
In the following constraint, for all $e$ and $t$, the fractional component controls the impact of each time step's information on the overall sum, with %larger values for more recent inspections and 
increasingly smaller values for less recent inspections.
% For all $e$ and $t$, we pose the following constraint.
\begin{align}
z_{e,t} \leq \sum_{t_h = \max{(t - \lambda, 1)}}^{t-1} \delta_{e,t_h} \frac{\lambda - (t - t_h) + 1}{\lambda (t - t_h)} \label{eq:inspection_decay}
\end{align}

Both $z_{e,t} \in [0,1]$ and $\delta_{e,t} \in \{0,1\}$ only contribute to the overall cost function (\ref{eq:overall_cost}) by reducing cost, so the optimizer will seek to maximize these values. Thus, equality will be achieved in both (\ref{eq:edge_inspected}) and (\ref{eq:inspection_decay}) up to their maximum bounds. 

\subsection{Uncertainty-Aware Multi-Robot Optimization Problem}
\label{sec:optimization_problem}
%%%%%%%%%%%%%%%%%%%%%%%%%%%%%%%%%%%%%%%%%%

% We express our overall MIP problem in Table~\ref{tab:optimization_problem}. We define each component of the optimization problem in the subsequent sections.

We combine our objective functions and constraints to form our overall MILP problem in Table~\ref{tab:optimization_problem},
% In this table, equation numbers starting with ``\cite{dimmig2023mip}-'' represent equation numbers from \cite{dimmig2023mip}, reproduced here for completeness. These include 
%cost constraints for overwatch opportunities, $C_{\Omega_{\scripto,t}}$, and 
% general constraints for setting support variables used in our cost functions and constraints to restrict movement to the topological graph. For the formulation of these constraints and further detail, please refer to \cite{dimmig2023mip}.
% We captured our costs and constraints linearly in a MILP, 
which we solve with the Gurobi optimizer \cite{GurobiOptimization2023}.

\begin{table}[tbh]
\centering
    \vspace*{2mm}
	\caption{MIP Optimization Problem with Uncertainty and Scouts}
    \vspace*{-8mm}
	\label{tab:optimization_problem}
	\begin{center}
		\renewcommand{\arraystretch}{2.0}
        \resizebox{\columnwidth}{!}{%
		\begin{tabular}{ p{0.01cm} p{6.0cm} p{1.7cm} | c }
			%	\hline
			%	\multicolumn{6}{|c|}{Agent} \\
			%	\hline
			%Variable Group Name & 
			\multicolumn{3}{c|}{\textbf{Optimization Problem}} & \textbf{Eq.} \\
			\hline
			\hline
			\multicolumn{3}{l|}{$\min~\sum\limits_{t = 1}^{n_T} \bigg( C_{T_t} + \sum\limits_{e \in E} \big( C_{W_{e,t}} + C_{U_{e,t}} \big) + \sum\limits_{v \in V} C_{F_{v,t}} \bigg)$ s.t.} & (\ref{eq:overall_cost})\\
            \hline
            \multirow{8}{*}{\rotatebox[origin=c]{90}{Scout and Uncertainty Constraints\quad\quad\quad}} 
            & $\breve{C}_{U_{A_{e,t}}} \geq \hat{u}_e (\phi_{e,t} - z_{e,t}),$ & $\forall e, t$ & (\ref{eq:robot_cost_uncertainty}) \\
            & $\breve{C}_{U_{K_{e,t}}} \geq \hat{u}_e (\theta_{e,t} - z_{e,t}),$ & $\forall e, t$ & (\ref{eq:scout_cost_uncertainty}) \\
            & $\theta_{e,t} \geq \frac{1}{n_K n_{\tau}} \sum\limits_{\tau = 1}^{n_{\tau}} q_{e,\tau,t},$ & $\forall e, t$ & (\ref{eq:edge_used})\\
            & $f_{v, t} \leq p_{v, t}, \quad q_{v, 1, t} = f_{v,t}, \quad q_{v, n_\tau, t} = f_{v, t},$ & $\forall v, t$ & (\ref{eq:scout_deploy}) \\
            % & $q_{v, 1, t} = f_{v,t}$ & $\forall v, t$ & (\ref{eq:scout_start_deploy}) \\ 
            % & $q_{s, \tau, 1} = n_s$ & $\forall s, \tau$ & (\ref{eq:scout_start_t1}) \\
            % & $q_{v, n_\tau, t} = f_{v, t}$ & $\forall v, t$ & (\ref{eq:scout_goal}) \\
            & $\sum_{l \in L} q_{l, \tau, t} = \sum_{v \in V} f_{v, t},$ & $\forall \tau, t$ & (\ref{eq:max_robots}) \\
            & $\sum\limits_{\substack{l_{ij} = (v_i, v_j) \in L}} q_{l_{ij},\tau-1, t} = \sum\limits_{\substack{l_{ji} = (v_j, v_i) \in L}} q_{l_{ji},\tau, t}, $ & $\forall t, v_j, \newline \tau \in [2, n_\tau]$ & (\ref{eq:scout_sequential}) \\
            % \hline
            % \multirow{6}{*}{\rotatebox[origin=c]{90}{Edge Inspection Constraints\quad\quad\quad}} 
            & $\delta_{e,t} \leq \sum_{\substack{\tau = 1}}^{n_\tau} {q_{e, \tau, t}},$ & $\forall e, \newline t \in [1, n_T - 2]$ & (\ref{eq:edge_inspected}) \\
            & $z_{e,t} \leq \sum_{t_h = \max{(t - \lambda, 1)}}^{t-1} \delta_{e,t_h} \frac{\lambda - (t - t_h) + 1}{\lambda (t - t_h)},$ & $\forall e, t$ & (\ref{eq:inspection_decay}) \\
            % & $\zeta_{e,t} \leq \phi_{e,t}$ & $\forall e, t$ & (\ref{eq:inspected_used_1}) \\ 
            % & $\zeta_{e,t} \leq z_{e,t}$ & $\forall e, t$ & (\ref{eq:inspected_used_2}) \\
            % & $\upsilon_{e,\tau,t} \leq \theta_{e,\tau,t}$ & $\forall e, \tau, t$ & (\ref{eq:inspected_used_scouts_1}) \\
            % & $\upsilon_{e,\tau,t} \leq z_{e,t}$ & $\forall e, \tau, t$ & (\ref{eq:inspected_used_scouts_2}) \\
            % \multirow{5}{*}{\rotatebox[origin=c]{90}{Cost Constraints~}} 
            % & $C_{W_{e,t}} \geq - m_e p_{e, t} + (\bar{w}_e + m_e a_e) \phi_{e,t} + \sum\limits_{\tau = 1}^{n_{\tau}} \bar{w}_e q_{e,\tau,t},$  &$\forall e, t$ & (\ref{eq:cost:trav1})\\
            % & $C_{W_{e,t}} \geq - r_e p_{e, t} + (\bar{w}_e + r_e a_e) \phi_{e,t} + \sum\limits_{\tau = 1}^{n_{\tau}} \bar{w}_e q_{e,\tau,t},$ &$\forall e, t$ & (\ref{eq:cost:trav2})\\
            % & $C_{\Omega_{\scripto,t}} \geq - \frac{\omega_\scripto}{\alpha_\scripto} p_{v_i, t},$ &$\forall \scripto, t $ & \cite{dimmig2023mip}-(8)\\
            % & $C_{\Omega_{\scripto,t}} \geq - \omega_\scripto - \gamma_\scripto (p_{v_i, t} - \alpha_\scripto),$ &$\forall \scripto, t $ & \cite{dimmig2023mip}-(9)\\
            % & $C_{\Omega_{\scripto,t}} \geq - \frac{\omega_\scripto}{\alpha_\scripto} n_A p_{e_j, t},$ &$\forall \scripto, t $ & \cite{dimmig2023mip}-(10)\\
            \hline
            \multirow{5}{*}{\rotatebox[origin=c]{90}{Carrier Robot Constraints\quad}} 
            & $\psi_t \geq \frac{1}{n_A + n_K n_{\tau}} \sum\limits_{e \in E} \bigg( p_{e,t} + \sum\limits_{\tau = 1}^{n_{\tau}} q_{e,\tau,t} \bigg),$ &$\forall t $ & (\ref{eq:constr:time_vars}) \\
            & $\phi_{e,t} \geq \frac{1}{n_A} p_{e,t},$ & $\forall e, t $ & (\ref{eq:edge_used}) \\ %\cite{dimmig2023mip}-(13) \\
            & $p_{s, 1} = n_{s}, \quad p_{d, n_T} \geq n_{d}$ &$\forall s, d $ & (\ref{eq:start_goal_constr}) \\ %\cite{dimmig2023mip}-(15)\\
            % & $p_{d, n_T} \geq n_{d},$ &$\forall d $ & (\ref{eq:goal_constr}) \\ %\cite{dimmig2023mip}-(16)\\
            & $\sum\limits_{l \in L} p_{l,t} = n_A,$ &$\forall t$ & (\ref{eq:max_robots}) \\%\cite{dimmig2023mip}-(17)\\
            & $\sum\limits_{l_{ij} = (v_i, v_j) \in L} p_{l_{ij},t-1} = \sum\limits_{l_{ji} = (v_j, v_i) \in L} p_{l_{ji},t},$ &$\forall v_j, \newline t \in [2, n_T]$ & (\ref{eq:carrier_sequential}) \\ %\cite{dimmig2023mip}-(18) \\
            \hline
		\end{tabular}%
        }
	\end{center}
    \vspace*{-6mm}
\end{table}

% %%%%%%%%%%%%%%%%%%%%%%%%%%%%%%%%%%%%%%%%%%
% \section{ENVIRONMENT SEGMENTATION}
% \label{sec:optimization_problem}
% %%%%%%%%%%%%%%%%%%%%%%%%%%%%%%%%%%%%%%%%%%

% As a representative test scenario, we consider reconnaissance in a meadow environment from \cite{NatureManufacture}. 
% To generate our dynamic topological graphs we...

% Uncertainty...

%%%%%%%%%%%%%%%%%%%%%%%%%%%%%%%%%%%%%%%%%%%%%%%%%%%%%%%%%%%%%%%%%%%%%%%%%%%%%%%%
\section{COMPUTATIONAL RESULTS AND DISCUSSION}

We evaluate our approach in scenarios that demonstrate the benefit of uncertainty reduction using scout robots. Scouts generally prioritize exploring high uncertainty edges and edges that are planned to be traversed to provide the greatest cost reduction. %; exploration of these edges provides the greatest cost reduction. 
In the following examples, we use demonstrative values for the parameters in Table~\ref{tab:parameters}. %, which can be adjusted for different objectives. 
For a particular scenario, parameters $\xi$, $\lambda$, and $\beta$, as well as any weights added to the cost functions, are set based on the desired risk tolerance, completeness of exploration, and/or frequency of exploration. In particular, $\xi$ controls the prioritization of exploring all edges versus edges planned to be traversed, smaller $\lambda$ values will trigger more frequently revisiting explored edges, and $\beta$ reflects the tolerance of risk (closer to $0$ being the most risk averse).  Unless noted otherwise, in the following examples we used $\xi = 1$, $\lambda = 5$, and $\beta = 0$. 
\tflag{R-1-3}\blue{With the exception of the ablation study, we construct graphs with randomly generated expected edge costs and uncertainty to evaluate our algorithm; %(with the exception of the ablation study), %where we use simple values), 
however, we note that these values could be generated from continuous space as in \cite{dimmig2024workshop}.}
%This approach can be applied to other scenarios 

\subsection{Ablation Study}

\begin{figure}[t]
	\centering
    \vspace*{2mm}
	% \begin{subfigure}[t]{0.9\columnwidth}
		\centering
		\includegraphics[width=\columnwidth]{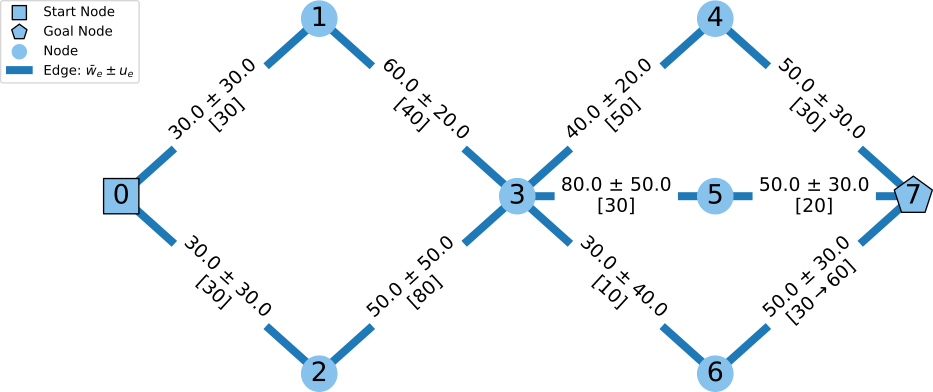}
		\captionsetup{font=scriptsize}
        \vspace*{-2mm}
        \caption{Example dynamic topological graph with expected edge weights, uncertainty, and the true cost (in brackets) labeled for each edge. The true cost of edge (6,7) changes over time from 30 to 60.}
	   \label{fig:toy102_true_cost}
	% \end{subfigure}
    % \vspace*{-1mm}
\end{figure}

To demonstrate the advantages of each component of our algorithm, we performed an ablation study on the example graph in Fig.~\ref{fig:toy102_true_cost}. We consider symmetric uncertainty about an expected value of each edge weight. %as denoted in Fig.~\ref{fig:toy102_true_cost}. %, and the true values of the edge weights are shown in brackets. 
The true values can be observed by the scouts %traversing the corresponding edges 
and evolve (e.g., due to the observer moving), so frequent exploration will yield updated values. %(e.g., the evolving cost on edge (6,7), such as due to the observer moving). 
% To represent this, on edge (6,7) the true cost starts at 30, but goes to 80 at later time steps. This could be due to an observer moving to more easily view (6,7). 
All robot units (carrier robots with scouts) start at node 0. The overall goal is for some subset of units to reach node 7 within 8 time steps $t$. The cost reduction for teaming, $r_e$, is $1$ on all edges to incentivize moving together (though reducing uncertainty is often more valuable). %, however this small cost reduction is easily overpowered by other priorities (such as reducing uncertainty). 
Scouts can move 8 scout time steps, $\tau$, within one time step, $t$. %, (i.e., 8 times faster than carrier robots). 
Scouts' movement costs a quarter of the cost of an edge (weight and uncertainty) (i.e., $\zeta = 0.25$), as they would be less detectable. 
%Scouts can launch at all time steps except for the last, since there is no value to the information gathered once the team has reached the goal. 
%
\begin{figure}[t]
    \centering
    \vspace*{-2mm}
	\begin{subfigure}[t]{0.48\columnwidth}
		\centering
		\includegraphics[width=\textwidth]{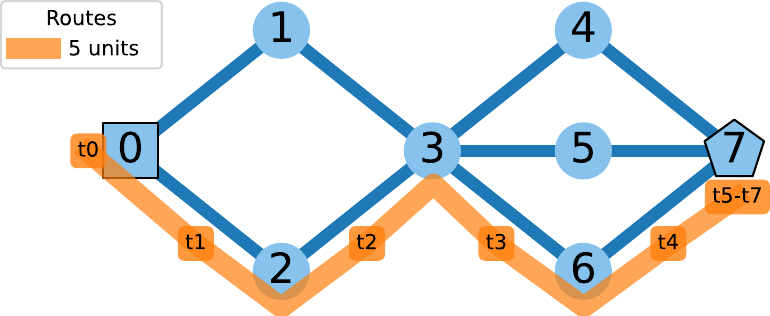}
		\captionsetup{font=scriptsize}
        \caption{Expected edge weights without uncertainty and scouts, true~cost~=~180}
		\label{fig:ablation_just_weights}
	\end{subfigure}
    \hspace*{1mm}
	\begin{subfigure}[t]{0.48\columnwidth}
		\centering
		\includegraphics[width=\textwidth]{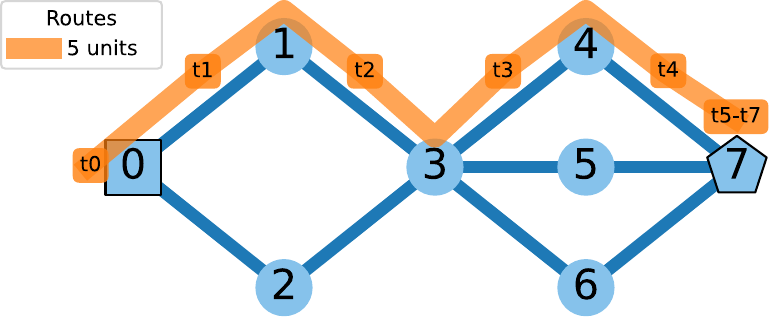}
		\captionsetup{font=scriptsize}
        \caption{Expected edge weights and uncertainty without scouts, true~cost~=~150}
		\label{fig:ablation_uncertainty}
	\end{subfigure}

	\begin{subfigure}[t]{0.48\columnwidth}
		\centering
		\includegraphics[width=\textwidth]{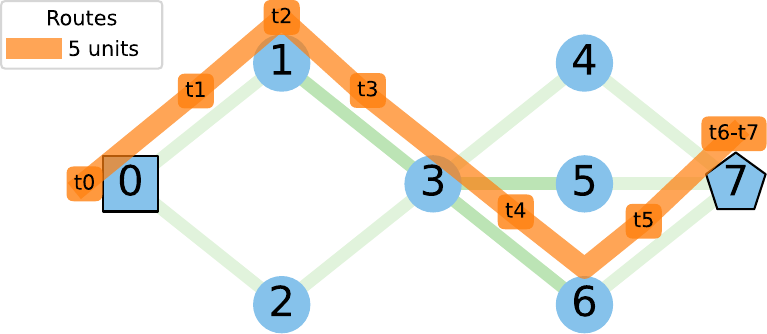}
		\captionsetup{font=scriptsize}
        \caption{Expected edge weights and uncertainty with scouts, without certainty decay, true~cost~=~140}
		\label{fig:ablation_scouts_wo_decay}
	\end{subfigure}
    \hspace*{1mm}
 	\begin{subfigure}[t]{0.48\columnwidth}
		\centering
		\includegraphics[width=\textwidth]{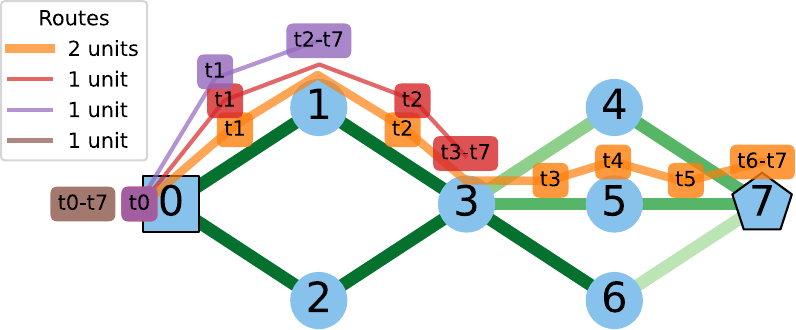}
		\captionsetup{font=scriptsize}
        \caption{Expected edge weights and uncertainty with scouts and certainty decay, true~cost~=~120}
		\label{fig:ablation_scouts_w_decay}
	\end{subfigure}
 
    \vspace*{-1mm}
    \begin{center}
	\begin{subfigure}[t]{0.68\columnwidth}
		\centering
        \includegraphics[width=\textwidth]{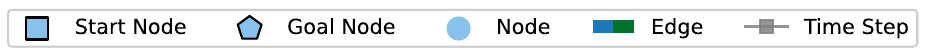}
    \end{subfigure}
    \begin{subfigure}[t]{0.73\columnwidth}
		\centering
		\includegraphics[width=\textwidth]{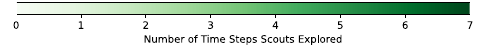}
	\end{subfigure}%
 \end{center}
    \vspace*{-5mm}
	\caption{Ablation study illustrating the advantages of considering uncertainty, scouts for the reduction of uncertainty, and certainty decay for evolving conditions. Each plot shows the final paths of the robot units through the graph. Each unit is composed of a carrier robot and scout. The first case is the baseline from \cite{dimmig2023mip} and each subsequent case adds one of the components from the algorithm proposed in this paper to show the resulting effect to the final team routes through the graph. 
    In (c) and (d), 
    % In the two cases that include scout robot deployments 
    the darkness of the edges reflect the number of time steps the scouts explored that edge.}
	\label{fig:ablation}
    \vspace*{-6mm}
\end{figure}
%
%%% Valuable clarification, but likely too verbose
% We consider four cases in our ablation study in Fig.~\ref{fig:ablation}: %Figures~\ref{fig:ablation_just_weights}-\ref{fig:ablation_scouts_w_decay} 
% %with varying information considered: 
% \begin{enumerate}[(a)]
%     \item Expected edge weights alone, as implemented in \cite{dimmig2023mip} \label{case:ablation_just_weights}
%     \item Expected edge weights with uncertainty (i.e., including cost (\ref{eq:total_cost_uncertainty}) and cost constraint (\ref{eq:robot_cost_uncertainty}), excluding scout terms)\label{case:ablation_uncertainty}
%     \item Uncertain edge weights with the ability to deploy scouts to reduce the uncertainty to zero (i.e., (\ref{eq:inspection_decay}) is modified to sum $\delta_{e,t_h}$ without any $\lambda$ term) \label{case:ablation_scouts_wo_decay}
%     \item Uncertain edge weights with the ability to deploy scouts with decay on the certainty gained from the scouts actions, as influenced by (\ref{eq:inspection_decay}) \label{case:ablation_scouts_w_decay}
% \end{enumerate}
% Each of these scenarios is solved for using our optimization problem, in Table~\ref{tab:optimization_problem}, with the corresponding components removed from the algorithm. The resulting final team paths for each are shown in Fig.~\ref{fig:ablation}, as well as the true cost for the route to the goal node. 
% For simplicity, we use a coefficient of optimism, $\beta$, of 0 to find solutions that are most risk averse.%; however $\beta$ could be set based on the risk profile.
Fig.~\ref{fig:ablation} depicts the final team paths and true cost for the route to the goal node for each ablation case where the noted elements were removed from our optimization problem in Table~\ref{tab:optimization_problem}.

%Since we are considering a reconnaissance test case and for simplicity in this demonstration, we use a coefficient of optimism, $\beta$, of 0 to find solutions that are most risk averse, but depending on the application space $\beta$ could be appropriately set.

\begin{figure}[t]
    \vspace*{2mm}
	\centering
	\begin{subfigure}[t]{0.48\columnwidth}
		\centering
		\includegraphics[width=\textwidth]{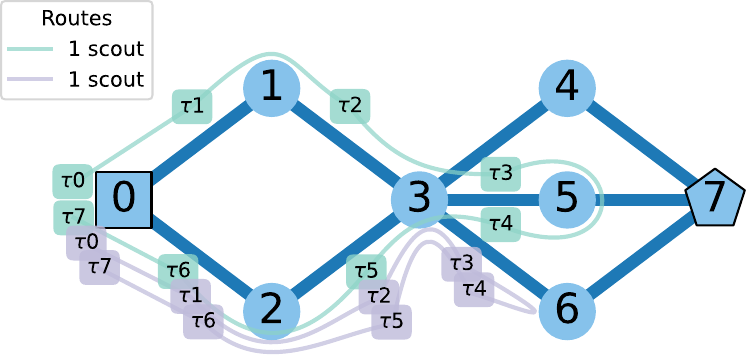}
		\captionsetup{font=scriptsize}
        \caption{Scout routes at t = 0}
	   \label{fig:scouts_wo_decay_0}
	\end{subfigure}
    %\vspace*{2mm}
    %
	\begin{subfigure}[t]{0.48\columnwidth}
		\centering
		\includegraphics[width=\textwidth]{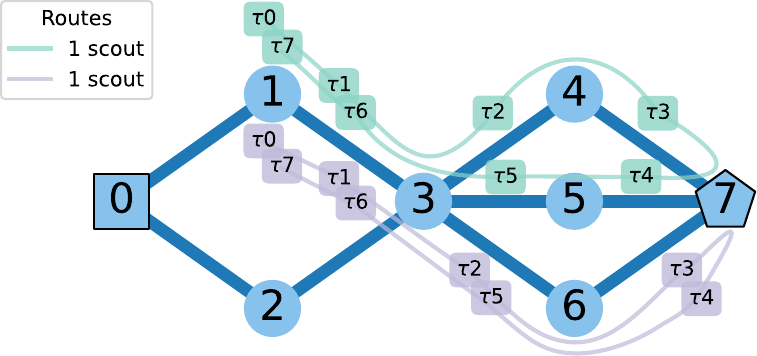}
		\captionsetup{font=scriptsize}
        \caption{Scout routes at t = 2}
		\label{fig:scouts_wo_decay_2}
	\end{subfigure}
    \vspace*{-2mm}
    \begin{center}
	\begin{subfigure}[t]{0.68\columnwidth}
		\centering
        \includegraphics[width=\textwidth]{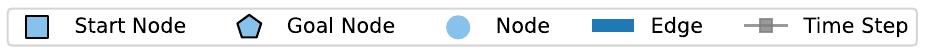}
	\end{subfigure}%
    \end{center}
    \vspace*{-6mm}
	\caption{Scout routes when deployed in the ablation case in Fig.~\ref{fig:ablation_scouts_wo_decay}. 
    %Full team routes are shown in Fig.~\ref{case:ablation_scouts_wo_decay}.
    }
	\label{fig:scout_paths}
    \vspace*{-3mm}
\end{figure}

In Fig.~\ref{fig:ablation_just_weights}, 
%case~\ref{case:ablation_just_weights}, 
the path with minimum expected edge cost is selected. 
% However, this path has very high uncertainty, so the resulting true cost is significantly higher than expected. 
However, the high uncertainty results in a high true cost.
When the uncertainty is considered in Fig.~\ref{fig:ablation_uncertainty},
%case~\ref{case:ablation_uncertainty}, 
a path that minimizes the expected weights and the maximum uncertainty is found. This plans for a worst-case scenario. %, but the resulting true cost is less than the expected cost. 
In Fig.~\ref{fig:ablation_scouts_wo_decay}, %case~\ref{case:ablation_scouts_wo_decay}, 
scouts are deployed at nodes 0 and 1 to reduce the uncertainty in planning. The scout paths are shown in Fig.~\ref{fig:scout_paths}. These deployments result in a more informed route, however, conditions are continuously evolving, and exploring each location once is not sufficient to see the future cost increase associated with traversing edge (6,7). Finally, we introduce decaying certainty to our algorithm in Fig.~\ref{fig:ablation_scouts_w_decay},
%case~\ref{case:ablation_scouts_w_decay}, 
such that uncertainty increases after a scout visit to incentivize revisiting locations for updated information. This results in a team plan 
%in Fig.~\ref{fig:ablation_scouts_w_decay} 
that minimizes the true cost. 
% We see robots stay behind at nodes to be able to deploy scouts to reduce the overall uncertainty; these routes look similar to those in Fig.~\ref{fig:scout_paths}. 
In this case, the benefit of frequent scout deployments outweighed traversing together, which 
resulted in units staying behind at nodes.
% ultimately breaking the team apart. 
This ablation study emphasizes the differences between each ablation case; however, the cost function can be weighted differently based on the priorities of a scenario. 
%This motivates adding scouts to gather more information to be able to plan a risk-averse path based on updated information.

%Our approach does not guarantee minimizing the true cost of traversing a path, as it did in this example, but provides metrics for planning to minimize the overall uncertainty and uncertainty of traversed paths to be able to plan multi-agent routes towards our goals of minimizing detectability and maximizing safety. 

\subsection{Coefficient of Optimism}
Our coefficient of optimism $\beta$ from the Hurwicz Criterion %, as employed in (\ref{eq:uncertain_weights}),
reflects our tolerance of risk. 
Setting $\beta = 0$ is the most risk averse approach and considers the uncertainty in the worst-case scenario.
% The most risk averse approach is setting $\beta = 0$, which considers the uncertainty in the worst-case scenario. 
This results in scouts exploring more to reduce the overall risk. Fig.~\ref{fig:beta} shows the frequency scouts explore each edge for various $\beta$ values. $\beta$ could be set based on the operational scenario and tolerance for risk. %In Fig.~\ref{fig:beta} the edge weights and uncertainty values are randomized.

\subsection{Combinatorial Considerations and Computation Time}
We investigated how the computation time of our algorithm scales with the number of decision variables when we apply our key innovations (removing the decision space dependence on the number of agents and the linear formulation of our cost functions and constraints). 
% We investigated how the algorithm from \cite{dimmig2023mip} scales for considering uncertainty and adding a new class of vehicles, scouts. 
Our total number of decision variables scales by 
%\begin{align}
% $n_T (1 + n_L + n_E + \mathbf{n_L n_{\boldsymbol\tau} + 5 n_E + n_V})$. %\label{eq:vars_scaling}
$n_T (1 + n_L + n_E + n_L n_{\tau} + 5 n_E + n_V)$. %\label{eq:vars_scaling}
%\end{align}
% Thus the most significant components are the number of locations ($n_L = n_E + n_V$) and the number of time steps for each type of vehicle ($n_T$ and $n_{\tau}$). 
% Here we have indicated the additions from the algorithm in \cite{dimmig2023mip} in bold. 
The greatest impact is due to the scout time step, $\tau$, occurring within each carrier robot time step, $t$. %, which adds $n_T n_L n_{\tau}$ variables to capture full scout trajectories at each $t$. 
% Ultimately, we were able to leverage the key innovation in \cite{dimmig2023mip} to formulate the problem such that the number of robots is removed from our decision space to keep the solve time tractable for larger graph sizes. 

\begin{figure}[t]
    \centering
    \vspace*{2mm}
	\begin{subfigure}[t]{0.23\columnwidth}
		\centering
		\includegraphics[width=\textwidth]{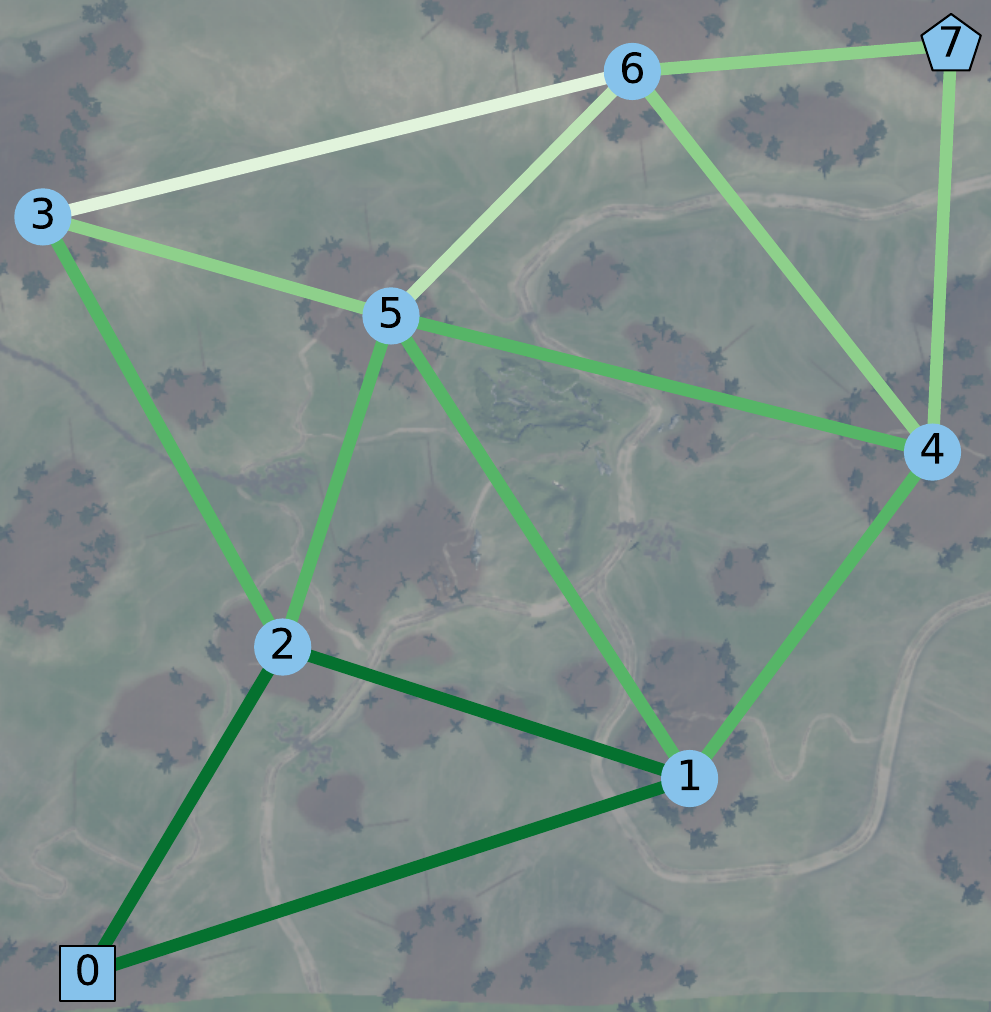}
		\captionsetup{font=scriptsize}
        \caption{$\beta = 0$}
		\label{fig:beta0}
	\end{subfigure}
    % \hspace*{1mm}
	\begin{subfigure}[t]{0.23\columnwidth}
		\centering
		\includegraphics[width=\textwidth]{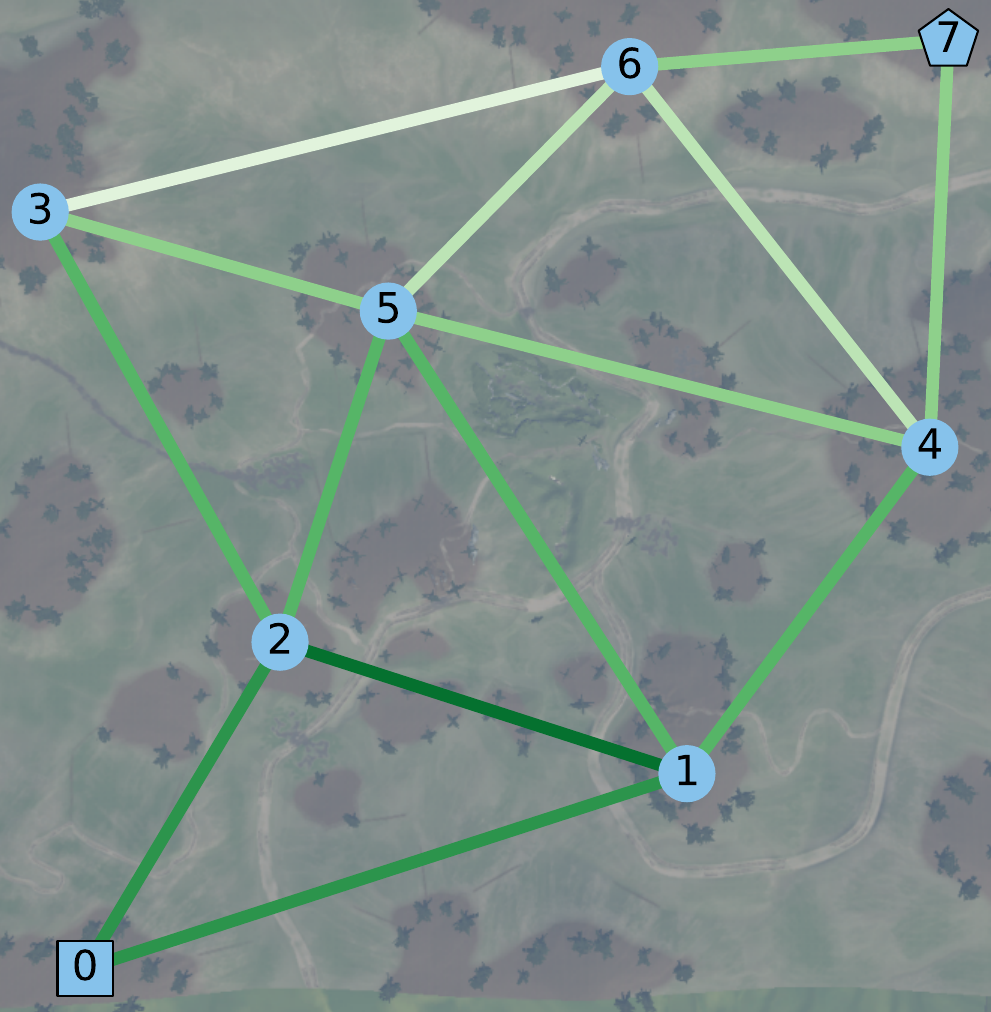}
		\captionsetup{font=scriptsize}
        \caption{$\beta = 0.15$}
		\label{fig:beta0-15}
	\end{subfigure}
    %
    % \vspace*{2mm}
	\begin{subfigure}[t]{0.23\columnwidth}
		\centering
		\includegraphics[width=\textwidth]{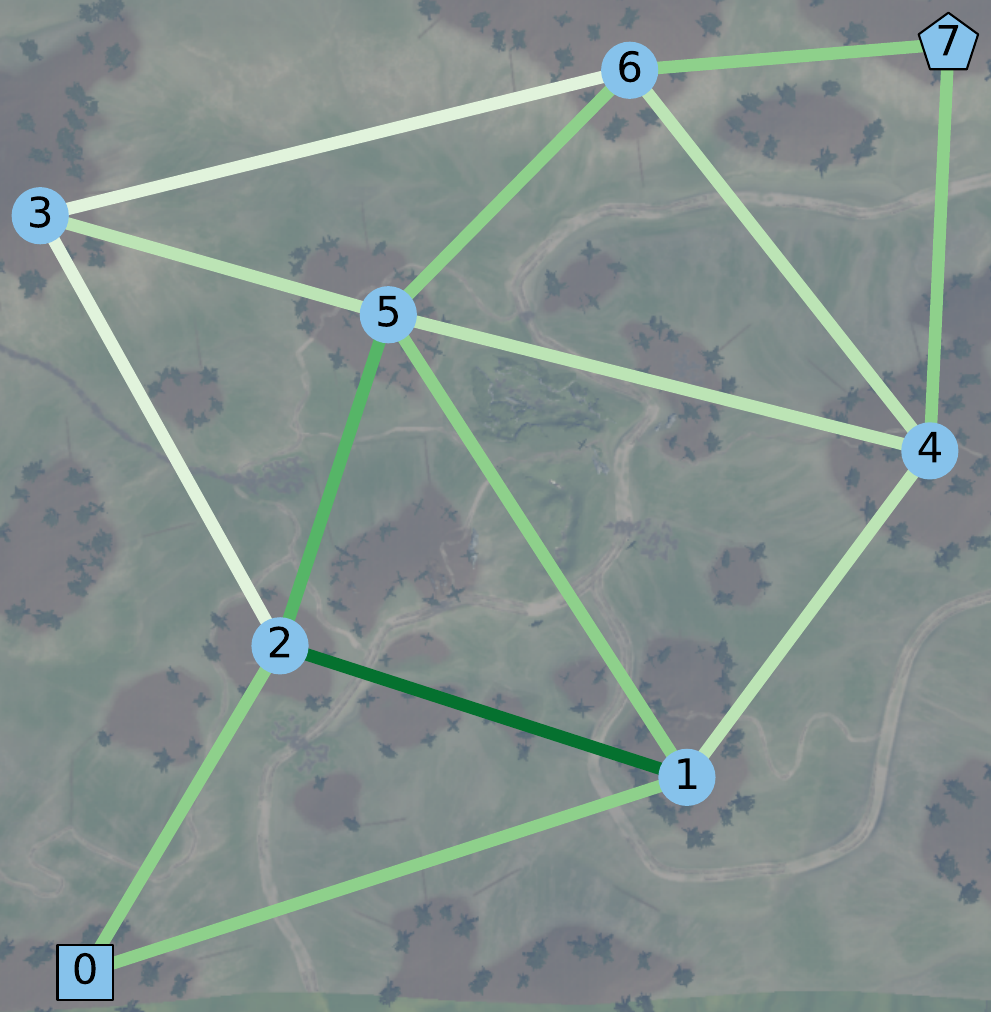}
		\captionsetup{font=scriptsize}
        \caption{$\beta = 0.30$}
		\label{fig:beta0-30}
	\end{subfigure}
    % \hspace*{1mm}
 	\begin{subfigure}[t]{0.23\columnwidth}
		\centering
		\includegraphics[width=\textwidth]{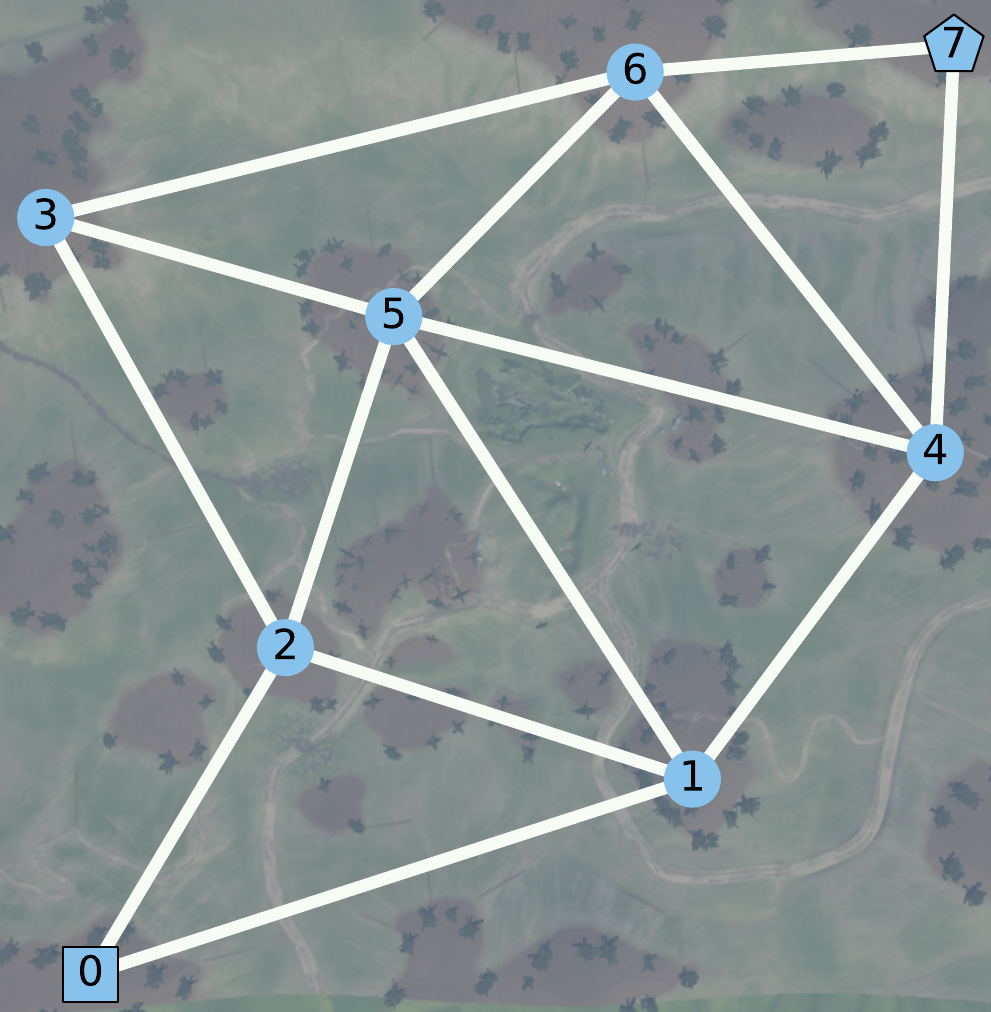}
		\captionsetup{font=scriptsize}
        \caption{$\beta = 0.45$}
		\label{fig:beta0-45}
	\end{subfigure}
    \vspace*{-2mm}
    \begin{center}
	\begin{subfigure}[t]{0.52\columnwidth}
		\centering
        \includegraphics[width=\textwidth]{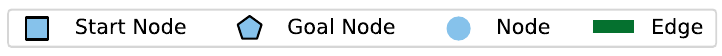}
	\end{subfigure}%
    \end{center}
    \vspace*{-4mm}
    \begin{center}
	\begin{subfigure}[t]{0.73\columnwidth}
		\centering
		\includegraphics[width=\textwidth]{figures/small_colorbar.pdf}
	\end{subfigure}%
    \end{center}
    \vspace*{-5mm}
	\caption{Scout exploration for different coefficients of optimism, $\beta$.}
	\label{fig:beta}
    \vspace*{-5mm}
\end{figure}

\begin{figure}[t]
	\vspace*{1mm}
	\centering
	\includegraphics[width=0.81\columnwidth]{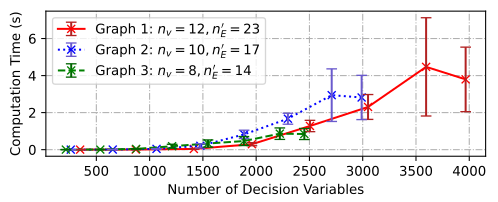}
    \vspace*{-4mm}
	\caption{Computation time for three example graphs of varying sizes versus the number of decision variables. Error bars are one standard deviation.
	}
	\label{fig:computation_time}
	\vspace*{-6mm}
\end{figure}

To adjust for new information and an evolving environment, we plan in a receding horizon. 
At each time step, we generate optimal solutions for the remainder of our time horizon after updating our graph with the information gathered by the scouts. 
% We first generate an optimal solution to our problem for the entire time horizon. After executing the first step in our solution, we update our graph with the information gathered by the scouts, set the new starting positions for each robot, and reduce the total number of time steps by 1. We then solve the optimization problem again and continue this process until the number of time steps goes to 1. 
\tflag{R-1-4}\blue{For real-time operation, we plan for the next step while the current step is executing. For these types of long distance plans ($\sim 100-250m$ edges), traversing one edge can take numerous minutes, so our computational requirement is to solve for a new plan in less time.}
To demonstrate the scaling of our approach, we plot the computation time for each iteration on three differently sized graphs versus the number of decision variables in Fig.~\ref{fig:computation_time}. Graph~1 is shown in Fig.~\ref{fig:meadow_example} and Graph~3 is shown in Fig.~\ref{fig:beta}. 
When planning through an environment, the data points in Fig.~\ref{fig:computation_time} occur right to left, with the total number of decision variables decreasing with the receding horizon.
% Fig.~\ref{fig:computation_time} shows the total computation decreases for each subsequent plan as the total number of variables decreases. 
We averaged computation times across 100 trials with randomized edge weights and uncertainty. We used the Gurobi optimizer \cite{GurobiOptimization2023} on an Intel® Core™ i7-10875H CPU @ 2.30GHz × 16. The longest computation time remains on the order of seconds, which enables regular re-planning. 
\section{CONCLUSION}

In this paper, we explored multi-robot planning on dynamic topological graphs using mixed-integer programming for two challenging cases: heterogeneous teams and planning under uncertainty. We use the uncertainty in our problem as a motivation for a heterogeneous team and introduce scout robots that can investigate the environment to reduce the uncertainty of future team actions. Our approach results in a MILP problem that can be solved rapidly with a receding horizon in real-world scenarios. We tested this approach in example scenarios and demonstrated the ability to successfully generate risk-aware plans for multi-robot teams.
% Future work?In specific, we presented a novel approach to modeling uncertainty across a dynamic topological graph and demonstrated scaling to heterogeneous teams. 


%%%%%%%%%%%%%%%%%%%%%%%%%%%%%%%%%%%%%%%%%%%%%%%%%%%%%%%%%%%%%%%%%%%%%%%%%%%%%%%%

%%%%%%%%%%%%%%%%%%%%%%%%%%%%%%%%%%%%%%%%%%%%%%%%%%%%%%%%%%%%%%%%%%%%%%%%%%%%%%%%

%%%%%%%%%%%%%%%%%%%%%%%%%%%%%%%%%%%%%%%%%%%%%%%%%%%%%%%%%%%%%%%%%%%%%%%%%%%%%%%%
%\section*{APPENDIX}

%%%%%%%%%%%%%%%%%%%%%%%%%%%%%%%%%%%%%%%%%%%%%%%%%%%%%%%%%%%%%%%%%%%%%%%%%%%%%%%%
\section*{ACKNOWLEDGMENT}

\blindreview{[Placeholder for acknowledgment]}
\original{We gratefully acknowledge the support of the Army Research Laboratory under grant W911NF-22-2-0241.}

%%%%%%%%%%%%%%%%%%%%%%%%%%%%%%%%%%%%%%%%%%%%%%%%%%%%%%%%%%%%%%%%%%%%%%%%%%%%%%%%

\bibliographystyle{IEEEtran}
\bibliography{references.bib}

\end{document}